\documentclass[11pt]{article}

\usepackage[a4paper,margin=1in]{geometry}
\usepackage[T1]{fontenc}
\usepackage{amsmath}
\usepackage{stix2}
\usepackage{microtype}
\usepackage{amsthm}
\usepackage{graphicx}
\usepackage{booktabs}
\usepackage[numbers]{natbib}
\usepackage{placeins}
\usepackage{float}
\usepackage{adjustbox}
\usepackage[font=small,labelfont=bf]{caption}
\usepackage{authblk}
\usepackage[hidelinks]{hyperref}

\providecommand{\sniDefVal}[2]{\expandafter\def\csname num@#1\endcsname{#2}}
\sniDefVal{asym.n64train.drift}{14.5}
\sniDefVal{asym.n64train.rate}{0.40}
\sniDefVal{asym.s14n16.allpair}{0.27}
\sniDefVal{asym.s14n16.rate}{0.15}
\sniDefVal{asym.s14n64.allpair}{6.77}
\sniDefVal{asym.s14n64.rate}{3.42}
\sniDefVal{blk.16.blocks}{1.05}
\sniDefVal{blk.16.bridges}{2.52}
\sniDefVal{blk.16.edges}{24.8}
\sniDefVal{blk.16.ge90}{32}
\sniDefVal{blk.16.share}{83.5}
\sniDefVal{blk.64.blocks}{1.00}
\sniDefVal{blk.64.share}{99.1}
\sniDefVal{blk.comp.exmax}{0.993}
\sniDefVal{blk.comp.exmin}{0.978}
\sniDefVal{blk.mask.exmax}{0.969}
\sniDefVal{blk.mask.exmin}{0.941}
\sniDefVal{blk.rel.exmax}{1.000}
\sniDefVal{blk.rel.exmin}{0.979}
\sniDefVal{blk.rel.highdef}{82/3{,}160}
\sniDefVal{blk.rel.highrate}{2.6}
\sniDefVal{blk.rel.lowdef}{1/135}
\sniDefVal{comp.drift.tf}{0.125}
\sniDefVal{comp.mpnn16.allpair}{0.01}
\sniDefVal{comp.mpnn32.allpair}{0.00}
\sniDefVal{comp.mpnn32.exact}{9{,}183}
\sniDefVal{comp.mpnn64.allpair}{1.48}
\sniDefVal{head.allpair}{75.7}
\sniDefVal{head.defects}{945}
\sniDefVal{head.exact}{1{,}522}
\sniDefVal{head.f1}{0.9815}
\sniDefVal{head.rate}{62.1}
\sniDefVal{head.seedmax}{65.9}
\sniDefVal{head.seedmin}{53.3}
\sniDefVal{inrange.max}{0.45}
\sniDefVal{inrange.min}{0.26}
\sniDefVal{mask.allmax}{8.9}
\sniDefVal{mask.allmin}{5.7}
\sniDefVal{mask.gdt.all.allpair}{6.98}
\sniDefVal{mask.gdt.attn.allpair}{11.07}
\sniDefVal{mask.gdt.attn.f1}{0.9816}
\sniDefVal{mask.inr.attn}{2.02}
\sniDefVal{mask.inr.rel}{0.87}
\sniDefVal{mask.spd.all.allpair}{8.89}
\sniDefVal{mask.spd.allpair}{83.99}
\sniDefVal{mask.spd.attn.allpair}{14.18}
\sniDefVal{mpnn.exact}{9{,}110}
\sniDefVal{mpnn.n64exact}{183}
\sniDefVal{mpnn.rate}{6.06}
\sniDefVal{mpnn32.bridge.rate}{6.21}
\sniDefVal{n64train.exact}{9{,}155}
\sniDefVal{n64train.f1}{0.9987}
\sniDefVal{n64train.rate}{16.08}
\sniDefVal{ood.erm.max}{0.964}
\sniDefVal{ood.erm.min}{0.960}
\sniDefVal{ood.trained.max}{0.942}
\sniDefVal{ood.trained.min}{0.761}
\sniDefVal{params.gdt}{15{,}090{,}112}
\sniDefVal{r2.erm.allpair}{2.91}
\sniDefVal{r2.erm.edit}{99.5}
\sniDefVal{r2.erm.f1}{0.9984}
\sniDefVal{r2.erm.portal}{97.6}
\sniDefVal{r2.erm.portal_c}{97.8}
\sniDefVal{r2.erm.select}{97.0}
\sniDefVal{r2.gdt.allpairmax}{2.02}
\sniDefVal{r2.gdt.allpairmin}{0.52}
\sniDefVal{r2.gdt.editmax}{99.6}
\sniDefVal{r2.gdt.editmin}{99.5}
\sniDefVal{r2.gdt.lowterc}{99.0}
\sniDefVal{r2.gdt.portalmax}{79.0}
\sniDefVal{r2.gdt.portalmin}{66.0}
\sniDefVal{r2.idf1.max}{0.9987}
\sniDefVal{r2.idf1.min}{0.9981}
\sniDefVal{r2.klo1.portal}{73.9}
\sniDefVal{r2.klo10.select}{98.9}
\sniDefVal{r2.lam0.portal}{76.9}
\sniDefVal{r2.lam10.portal_c}{68.9}
\sniDefVal{r2.lam10.select}{98.9}
\sniDefVal{r2.merm.allpair}{8.30}
\sniDefVal{r2.merm.edit}{97.7}
\sniDefVal{r2.merm.f1}{0.9909}
\sniDefVal{r2.merm.portal}{82.5}
\sniDefVal{r2.mlam0.allpair}{9.17}
\sniDefVal{r2.mlam0.edit}{98.1}
\sniDefVal{r2.mlam0.portal}{78.9}
\sniDefVal{r2.mlam1.allpair}{7.29}
\sniDefVal{r2.mlam1.edit}{97.8}
\sniDefVal{r2.mlam1.portal}{75.1}
\sniDefVal{r2.mlam10.allpair}{7.26}
\sniDefVal{r2.mlam10.edit}{95.1}
\sniDefVal{r2.mlam10.f1}{0.9795}
\sniDefVal{r2.mlam10.portal}{64.4}
\sniDefVal{r2.paired.maxshare}{0.24}
\sniDefVal{r2.portal.pairs}{1{,}000}
\sniDefVal{r2.rel1.allpair}{2.02}
\sniDefVal{r2.rel1.portal}{79.0}
\sniDefVal{r2c.erm.allpair}{0.00}
\sniDefVal{r2c.erm.edit}{99.5}
\sniDefVal{r2c.erm.f1}{0.9986}
\sniDefVal{r2c.erm.portal}{98.1}
\sniDefVal{r2c.mpnn.badcells}{4}
\sniDefVal{r2c.mpnn.cells}{12}
\sniDefVal{r2c.mpnn.driftmax}{0.15}
\sniDefVal{r2c.mpnn.driftmin}{0.03}
\sniDefVal{r2c.mpnn.maxdef}{2}
\sniDefVal{r2p.blk.aug.gainmax}{7.6}
\sniDefVal{r2p.blk.aug.gainmin}{2.8}
\sniDefVal{r2p.blk.aug.pmax}{0.022}
\sniDefVal{r2p.blk.erm.pmin}{0.19}
\sniDefVal{r2p.gdt.onlyerm.max}{354}
\sniDefVal{r2p.gdt.onlyerm.min}{179}
\sniDefVal{r2p.gdt.onlytr.max}{16}
\sniDefVal{r2p.gdt.onlytr.min}{6}
\sniDefVal{r2p.gdt.pexp}{-36}
\sniDefVal{r2p.mlam0.nsig}{one}
\sniDefVal{r2p.mlam1.nsig}{all three}
\sniDefVal{r2p.moved}{4}
\sniDefVal{r2p.moved.maxdiff}{2}
\sniDefVal{r2p.mpnncells}{24}
\sniDefVal{r64.erm.allpair}{76.2}
\sniDefVal{r64.erm.edit}{95.3}
\sniDefVal{r64.erm.portal}{69.7}
\sniDefVal{r64.lam10.edit}{84.0}
\sniDefVal{r64.lam10.portal}{33.9}
\sniDefVal{r64.trained.allpairmax}{65.4}
\sniDefVal{r64.trained.allpairmin}{53.8}
\sniDefVal{r64.trained.exactmax}{121}
\sniDefVal{r64.trained.exactmin}{0}
\sniDefVal{repro.f1}{0.9634}
\sniDefVal{repro.f1max}{0.9671}
\sniDefVal{repro.f1min}{0.9624}
\sniDefVal{size.amp}{5.5}
\sniDefVal{size.edges16}{24.8}
\sniDefVal{size.edges64}{201.9}
\sniDefVal{size.pe16}{0.229}
\sniDefVal{size.pe64}{0.202}
\sniDefVal{size.peood}{1.263}
\sniDefVal{spd.defects}{1{,}013}
\sniDefVal{spd.exact}{1{,}338}
\sniDefVal{spd.rate}{75.7}
\sniDefVal{sssp.corrmax}{94.7}
\sniDefVal{sssp.corrmin}{86.8}
\sniDefVal{sssp.globmax}{6.6}
\sniDefVal{sssp.globmin}{6.5}
\sniDefVal{sssp.respmax}{98.0}
\sniDefVal{sssp.respmin}{93.6}
\sniDefVal{train.max.min}{34}
\sniDefVal{train.min.min}{6}
\sniDefVal{train.n64.hours}{7.6}
\sniDefVal{union.gdt.allpair}{75.8}
\sniDefVal{union.gdt.rate}{61.8}
\sniDefVal{union.mpnn.exact}{9{,}090}
\sniDefVal{union.mpnn32.exact}{9{,}103}
\sniDefVal{union.spd.defects}{1{,}028}
\sniDefVal{union.spd.exact}{1{,}383}

\makeatletter
\newcommand{\val}[1]{\@ifundefined{num@#1}%
  {\PackageError{numbers}{Undefined number `#1'}{}}%
  {\csname num@#1\endcsname}}

\makeatother

\newtheorem{proposition}{Proposition}

\theoremstyle{definition}
\newtheorem{definition}{Definition}
\theoremstyle{plain}

\title{Evaluating invariance training with certified answer-changing controls in neural graph algorithms}
\author[1]{Zhe Dong\thanks{Corresponding author: \texttt{zhe.dong@maine.edu}}}
\author[2]{Fang Qin\thanks{\texttt{fangq@stanford.edu}}}
\affil[1]{University of Maine at Presque Isle, Presque Isle, ME, USA}
\affil[2]{Stanford University, Stanford, CA, USA}
\date{}

\begin{document}
\maketitle

\begin{abstract}
Invariance training aims to make neural algorithmic reasoners robust to
answer-preserving input changes and is judged by accuracy and by how rarely such
edits change predictions. We ask whether this evaluation detects lost
sensitivity to answer-changing edits on minimum spanning trees of synthetic graphs, where both
directions can be certified. Attaching a subgraph through one bridge
provably preserves every original label. Attaching it through two edges,
oracle-selected to change a label, and lowering one weight inside the graph serve
as answer-changing controls; a response is a changed prediction on an edge whose
label changes. A reproduced graph transformer (GDT) within its published
accuracy range changes a provably fixed label on
\val{head.seedmin}--\val{head.seedmax}\% of the 64-vertex graphs it solves
exactly after a 64-vertex bridge attachment; removing the connecting edge
changes little. With 16-vertex graphs and attachments, three invariance
objectives lower GDT violations from \val{r2.erm.allpair}\% to
\val{r2.gdt.allpairmin}--\val{r2.gdt.allpairmax}\% of pairs at essentially
unchanged accuracy. Training also lowers the response to two-edge
attachments from \val{r2.erm.portal}\% to
\val{r2.gdt.portalmin}--\val{r2.gdt.portalmax}\%, while the response to in-base
weight edits stays near \val{r2.erm.edit}\%. At 64 vertices accuracy also falls,
but on 16-vertex bases only the attachment control reveals the loss. A
message-passing network loses some attachment response at small penalty weights but gains
little invariance. Running the conventionally trained GDT on each biconnected
block separately removes all observed violations and keeps accuracy and both
responses, so the task does not force the trade-off.
Sensitivity checks for trained invariance should include certified
answer-changing edits that attach material as the training edits do.
\end{abstract}

\section{Introduction}
\label{sec:intro}

Neural algorithmic reasoning trains neural networks to reproduce the outputs of
classical algorithms \citep{velickovic2020neural,velickovic2021nar,
velickovic2022clrs}. In many of these tasks, some changes to the input provably
leave the output on the original input unchanged: adding vertices that cannot be
reached from the source, for instance, leaves the shortest-path distance of
every original vertex as it was. A natural way to make a learned reasoner more
robust is therefore to train it to be invariant to such answer-preserving edits.
Hint-ReLIC \citep{bevilacqua2023hintrelic} does this for minimum spanning trees,
shortest paths and other algorithms by adding subgraphs that cannot change the
result, and \citet{rodionov2023nohints} regularise for invariance without
supervising intermediate algorithmic steps. Such methods are judged mainly by
accuracy on unedited inputs and, when invariance is measured, by how often
predictions on the original input change under answer-preserving edits.

These two measurements cannot tell a model that has learned which parts of the
input matter from one that has learned to disregard a whole class of edits.
Consider an accurate network that ignores every subgraph added to its input. Its
predictions never change under answer-preserving additions, and its accuracy on
unedited inputs is unaffected, so it scores well on both measures. When an added
subgraph does change the answer, however, it keeps the old answer, which is now
wrong. \citet{tramer2020tradeoffs} show that this happens in practice: training
for robustness within a perturbation set also makes models invariant to
perturbations in that set that change the true label. To check whether a model
draws the distinction, behavioural tests in natural language processing pair
edits that should not change a prediction with edits that should
\citep{niu2018oversensitivity,ribeiro2020checklist,gardner2020contrast}.

Two difficulties limit such tests. Whether an edit changes the correct answer
usually rests on human judgement. And the sensitivity test may use edits of a
different kind from those the model was trained to ignore. A model that has
learned to ignore only the trained kind of edit still responds to edits of other
kinds, so a test built from them reports no loss. This paper asks whether the second difficulty matters in
practice: can the choice of answer-changing control reverse the verdict on
invariance training?

\begin{figure}[t]
\centering
\includegraphics[width=\textwidth]{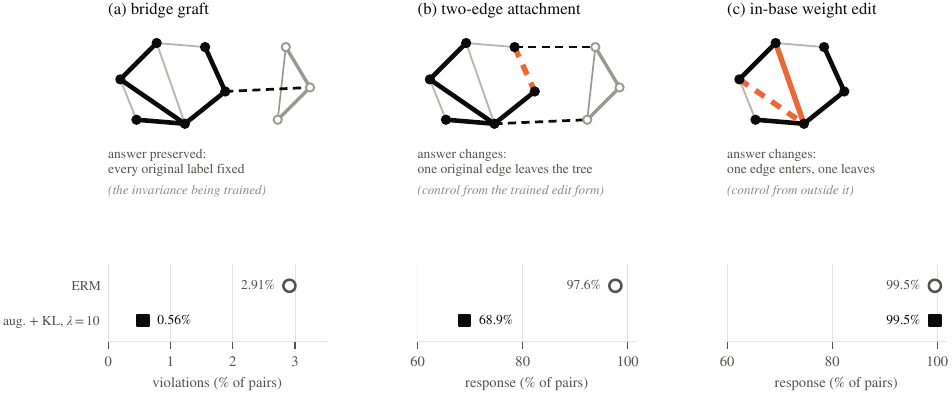}
\caption{The three certified edits, drawn on a schematic graph. Thick black
edges are in the minimum spanning tree, thin light-grey edges are the base's
other edges, darker grey edges and hollow vertices are attached material, dashed black lines are connecting edges, and orange marks the
labels that change (dashed when the edge leaves the tree). Bottom: the graph
transformer studied here, on 16-vertex graphs with 16-vertex attachments, after
standard training (ERM, open circle) and after invariance training (augmentation
with a KL penalty, $\lambda=10$, filled square); three-seed means. A violation is
a changed prediction on some original edge under (a); a response is a changed
prediction on an edge whose label the oracle changes under (b) or (c)
(Section~\ref{sec:res-training}).}
\label{fig:teaser}
\end{figure}

Graph algorithms remove the first difficulty and make the second one testable.
Consider the minimum spanning forest of a weighted graph $G$, and label each
edge by whether it belongs to the forest (Figure~\ref{fig:teaser}). If a connected subgraph $H$ is attached to $G$
through a single new edge, that edge is a bridge: no cycle passes through it, so
every original edge keeps its label (Proposition~\ref{prop:scope}). We call this
edit a \emph{bridge graft}. If the same kind of subgraph is attached through two
edges instead, a cycle through old and new material appears and an original edge
can leave the tree. An independent oracle certifies when this happens, and we
prove that exactly one original edge then changes label
(Proposition~\ref{prop:oneflip}). Finally, lowering the weight of a non-tree
edge inside $G$ far enough to bring it into the forest swaps exactly two labels,
that edge entering and one tree edge leaving, and attaches nothing
(Proposition~\ref{prop:exchange}). The bridge graft is the answer-preserving edit
that invariance training uses. The two-edge attachment is an answer-changing
control of the same form: like the graft, it adds a generated subgraph
(Section~\ref{sec:change}). The
in-base weight edit is an answer-changing control of a different form.
Single-source shortest paths admit the same construction, which we use only for
the structural experiment of Section~\ref{sec:res-sssp}.

We use these edits to evaluate a reproduction of the graph transformer of
\citet{gdt2025} (GDT), whose accuracy on this task lies within the published
range, and a message-passing network (MPNN). As in the original
work, the GDT is trained on 16-vertex graphs and tested on 64-vertex graphs.
Without invariance training it often violates the bridge-graft invariance: on
64-vertex graphs that it solves exactly, attaching a 64-vertex subgraph through a
bridge changes at least one original label on
\val{head.seedmin}--\val{head.seedmax}\% of graphs across three training seeds.
Adding the same subgraph with no connecting edge gives a similar rate, so at this
size the model is unstable to added material in general. Violations persist, at
much lower rates, when base and attachment together have no more vertices than a
training graph.

Invariance training reduces these violations, but the two answer-changing
controls disagree on what it costs. We train the GDT with three invariance
objectives and evaluate it on 16-vertex base graphs, the size of the training
graphs, with 16-vertex attachments. Every objective lowers the violation rate from \val{r2.erm.allpair}\% of pairs to
\val{r2.gdt.allpairmin}--\val{r2.gdt.allpairmax}\% at essentially unchanged
accuracy on unedited graphs. For the same models, the response to the two-edge attachment
falls from \val{r2.erm.portal}\% of pairs to
\val{r2.gdt.portalmin}--\val{r2.gdt.portalmax}\%, while the response to the
in-base weight edit stays near \val{r2.erm.edit}\%. An evaluation that relied on
the in-base edit would conclude that training improved invariance at no cost. On
64-vertex graphs, training also lowers accuracy, so some cost is visible anyway;
on 16-vertex bases accuracy is essentially unchanged, and only the control of the trained
form reveals the lower response.

Is this trade-off forced by the task? A bridge graft adds biconnected blocks,
the maximal pieces of a graph that stay connected when any single vertex is
deleted, without changing the existing ones. Running the trained network on each
block separately therefore makes the invariance hold by construction. On the GDT trained without an invariance
objective, this blockwise execution removes every observed violation while
keeping accuracy and both responses at their previous levels, so the task
permits the invariance without lowering either response. A blockwise predictor
trained for shortest paths likewise has no violations. Blockwise execution needs
the task's decomposition, so we use it as a reference, not as a general method.

Section~\ref{sec:related} reviews related work, Sections~\ref{sec:protocol}
and~\ref{sec:design} define the edits, metrics and experiments,
Section~\ref{sec:results} reports the results and Section~\ref{sec:discussion}
discusses what they imply for evaluating invariance training.

\section{Related work}
\label{sec:related}

\paragraph{Neural algorithmic reasoning and invariance}
Graph neural networks can be trained to execute classical graph algorithms step
by step \citep{velickovic2020neural}, and architectures whose computation aligns
with an algorithm's structure learn it more easily
\citep{xu2020reason,dudzik2022dp}. The CLRS benchmark collects such tasks
\citep{velickovic2022clrs}. Generalising beyond the training size depends on how
well the architecture and features match the task \citep{xu2021extrapolate} and
remains difficult on these benchmarks \citep{mahdavi2023ood}.
\citet{mahdavi2023ood} argue that standard augmentations such as Mixup and
DropEdge do not help here, because the labels of algorithmic tasks are not
preserved under them. Hint-ReLIC \citep{bevilacqua2023hintrelic} instead adds
material that cannot change the result; for graph algorithms such as
minimum spanning tree and shortest paths, this is a subgraph with no edge to the
original graph. Our bridge graft differs only in connecting the subgraph by one
edge, which still preserves every original label. Hint-ReLIC imposes invariance
on the representations of intermediate algorithmic steps, using the ReLIC
objective of \citet{mitrovic2021relic}, and \citet{rodionov2023nohints}
regularise for invariance without such steps. Robustness to irrelevant input
matters beyond algorithms as well: graph neural networks use graph structure even
when the target does not depend on it \citep{bechlerspeicher2024graphs}, and
language models are distracted by irrelevant context \citep{shi2023distracted},
which training on a mix of relevant and irrelevant retrieved passages reduces
\citep{yoran2024robust}. These works evaluate accuracy. We ask whether such
training preserves the response to subgraph additions that do change the answer.

\paragraph{Invariance bought by discarding information}
Invariance can be obtained by discarding information the task needs. In
invariant risk minimisation the null representation
satisfies the invariance penalty trivially and is excluded only by the risk term
\citep{arjovsky2019irm}, and classifiers can be invariant to many task-relevant
changes \citep{jacobsen2019excessive}. An invariance is useful only until the
transformation changes the label, as when a rotated 6 becomes a 9, which is
where learned augmentation ranges should stop \citep{benton2020augerino}.
Training for invariance has measurable costs: contrastive objectives harm tasks
that need the attribute their augmentations remove \citep{xiao2021contrastive},
and augmentation can lower accuracy on individual classes while raising it on
average \citep{kirichenko2023classlevel}. The closest result to ours is that of
\citet{tramer2020tradeoffs}: models trained to be robust within a norm ball are
also invariant to perturbations inside the ball that change the true label, as
judged by human annotators. Shortcut learning describes the broader pattern of
models satisfying a training objective by a rule that does not transfer
\citep{geirhos2020shortcut}.

\paragraph{Testing invariance and sensitivity together}
Behavioural tests in natural language processing check both invariance and
directional expectations \citep{ribeiro2020checklist}, contrast sets apply small
label-changing edits \citep{gardner2020contrast}, and
\citet{niu2018oversensitivity} pair should-not-change with should-change
perturbations, calling insensitivity to the latter over-stability.
Counterfactually augmented data trains on minimal label-changing pairs
\citep{kaushik2020cad,teney2020counterfactual}, and stress tests formalise
invariance to irrelevant features \citep{veitch2021counterfactual}. In each of
these works, whether an edited input should change the answer is decided by
people or by heuristics. Here both directions are certified, by a theorem for the
answer-preserving edit and by an oracle for the answer-changing ones, and the
two answer-changing controls differ in whether they attach material as the
trained edits do.

\paragraph{Graph decomposition}
Blocks, cut vertices and related decompositions are used in graph learning
\citep{zhang2023biconnectivity,dimitrov2023plane,chen2026cosparsify}, and
attention masking in graph transformers \citep{li2025damgt}. We use the classical
block decomposition, computed by depth-first search \citep{tarjan1972}, for a
different purpose: the network is run
on each block separately, so that a prediction depends only on its own block, and
we measure what this restriction costs in accuracy and sensitivity.

\paragraph{Guarantees for graph models}
Certified robustness to graph modification guarantees an unchanged prediction
when the number of modified nodes, edges or features stays below a threshold
\citep{li2025agnncert}; we instead ask for exact invariance under an unbounded
but restricted family of edits. Discrete neural algorithmic reasoners can be
proved correct on any test input \citep{rodionov2025discrete}, and graph neural
networks can provably learn some distributed graph algorithms exactly, with high
probability, on bounded-degree graphs \citep{fetrat2026exact}. An exactly correct
executor is trivially invariant to answer-preserving edits, so our test is aimed
at learned models with high empirical accuracy but no guarantee of correctness
on every input.

\section{Certified edits and evaluation protocol}
\label{sec:protocol}

\subsection{Tasks}

Let $G=(V,E,w)$ be an undirected graph with positive, pairwise distinct edge
weights. The first task is the minimum spanning forest (MST; on a connected graph, the
minimum spanning tree): the output is a label
$y_e\in\{0,1\}$ for every edge, equal to $1$ if $e$ belongs to the forest.
Because the weights are distinct, the forest is unique. The second task is
single-source shortest path (SSSP) from a fixed root $r$: the output for each vertex $v$
is its predecessor $\pi_v$, the vertex before $v$ on a shortest path from $r$,
and $d_G(v)$ denotes the distance from $r$ to $v$. Our models predict a distance
for every vertex, and predecessors are decoded from these predictions with a
fixed tie-breaking rule, needed because distinct weights do not make shortest
paths unique (Appendix~\ref{app:proofs}). All graphs come from the generator released with our primary
backbone \citep{gdt2025}. It links the components of an Erd\H{o}s--R\'enyi graph
but does not enforce connectivity; nearly all generated graphs are connected, and
the forest formulation covers the rest.

\subsection{The answer-preserving edit}
\label{sec:keep}

\begin{definition}[Bridge graft]
\label{def:graft}
Let $H=(V_H,E_H)$ be a connected weighted graph, vertex-disjoint from $G$. The
\emph{bridge graft} of $H$ onto $G$ at $a\in V$ and $h\in V_H$ is
$G'=(V\cup V_H,\,E\cup E_H\cup\{b\})$ with the new edge $b=\{a,h\}$, such that
all weights of $G'$ are positive and distinct.
\end{definition}

Because $b$ is the only edge between $V$ and $V_H$, it is a bridge of $G'$ and
lies on no cycle. Every cycle of $G'$ therefore lies inside $G$ or inside $H$.

\begin{proposition}[Protected outputs; classical]
\label{prop:scope}
For every bridge graft $G'$ of $G$: (i) each $e\in E$ lies in the minimum
spanning forest of $G'$ if and only if it lies in that of $G$; (ii) if the root
lies in $V$, then $d_{G'}(v)=d_G(v)$ for every $v\in V$ and the set of shortest
paths within $G$ is unchanged.
\end{proposition}

Part (i) follows from the cycle property: with distinct weights the minimum
spanning forest is unique \citep{kruskal1956}, an edge is left out of it exactly
when it is the heaviest edge on some cycle, and the graft creates no new cycle
through an original edge. Part (ii) holds because a path that leaves $V$ must cross $b$
twice, which cannot shorten it under positive weights.
Appendix~\ref{app:proofs} gives the proofs and shows that the tie-breaking rule
keeps decoded predecessors unchanged as well. For the model, however, the input
changes substantially: the number of tokens, the vertices reachable from the
root and the execution trace of both reference algorithms all change.

We call the outputs covered by Proposition~\ref{prop:scope}, the labels of
original edges and the predecessors of original vertices, the \emph{protected
outputs}. A pair $(G,G')$ is a \emph{violation} for a model when the model's
prediction on some protected output differs between $G$ and $G'$. The definition
uses discrete predictions only and involves no tolerance; changes in continuous
scores are reported separately as drift.

As a structural control we also use the \emph{disjoint union}, which places $H$
next to $G$ with no connecting edge. It preserves the answer trivially, and
comparing it with the bridge graft separates the effect of the added material
from that of the connecting edge.

\subsection{The answer-changing controls}
\label{sec:change}

A model that ignores every attached subgraph satisfies
Proposition~\ref{prop:scope} perfectly, so we pair the bridge graft with two
oracle-certified edits that change the answer. Both are defined for minimum
spanning forests; shortest paths use only the first
(Section~\ref{sec:metrics}). We use
\emph{attachment} as an operational term for any edit that adds a generated
subgraph to the base graph: the bridge graft and the first control below are
attachments, and the second control attaches nothing.

\paragraph{Two-edge attachment}
This control connects $H$ to $G$ through two edges $p_1=\{a_1,h_1\}$ and
$p_2=\{a_2,h_2\}$, with $a_1\neq a_2$ in $V$ and $h_1\neq h_2$ in $V_H$. Together
with paths inside $G$ and $H$, the two edges close a cycle through old and new
material, so an original edge can leave the forest. The whole pair, base graph,
subgraph and connecting edges, is redrawn until an independent Kruskal
implementation confirms that at least one original label changes.

The subgraph $H$ comes from the same generator as in the bridge graft, but the
edit differs from the graft in three other respects. It has two connecting edges
instead of one. Their weights are drawn from $(0,1)$ rather than from the
generator's range $(0,10)$, so that a heavy original edge is evicted more often.
And the redrawing retains only pairs in which such an eviction occurs, so the
two-edge pairs are sampled separately from the bridge-graft pairs.

\begin{proposition}[One original edge changes]
\label{prop:oneflip}
Let $G$ and $H$ be connected, let all weights of the two-edge attachment
$G'$ be distinct, and suppose some original label changes. Then exactly one edge
of $G$ changes label, and it changes from $1$ (in the forest) to $0$.
\end{proposition}

The proof (Appendix~\ref{app:proofs}) shows that the forest of $G'$ consists of
the two original forests and the two new edges, minus the heaviest edge of the
single cycle they form. For this control there is therefore exactly one affected
original edge, and the oracle identifies it.

\paragraph{In-base weight edit}
This control changes the answer without attaching anything. Choose a non-tree
edge $e$ of $G$ and let $f$ be the heaviest edge on the tree path between the
endpoints of $e$. Replace the weight of $e$ by a value drawn uniformly from
$(0,w_f)$ that collides with no existing weight.

\begin{proposition}[Exchange]
\label{prop:exchange}
After the edit, $e$ belongs to the minimum spanning forest, $f$ does not, and
every other label is unchanged.
\end{proposition}

This is the exchange property of spanning trees (Appendix~\ref{app:proofs}), and
the Kruskal oracle checks every edit as well. Unlike the two-edge attachments,
the in-base edits are built on the same base graphs as the bridge grafts, so
invariance and response can be measured on one sample
(Section~\ref{sec:metrics}).

\subsection{Metrics}
\label{sec:metrics}

A model's edge scores are decoded into a maximum-score spanning forest with
deterministic tie-breaking, so every MST prediction is a valid forest. Let $f(G)$
denote the decoded predictions on the protected outputs $P(G)$, and $E_f(G)$ the
event that all of them are correct; we then say that the model \emph{solves $G$
exactly}. Accuracy on unedited graphs is reported as \emph{structured
macro-F1}, the macro-F1 of the decoded forest's edge labels; where we compare
with the published accuracy of the backbone, we use its own metric,
\emph{per-edge macro-F1} on undecoded predictions. For
answer-preserving pairs we report two violation rates,
\begin{equation}
\label{eq:rates}
D_{\mathrm{all}}=\Pr\bigl[f(G')_{P(G)}\neq f(G)\bigr],\qquad
D_{\mathrm{exact}}=\Pr\bigl[f(G')_{P(G)}\neq f(G)\,\big|\,E_f(G)\bigr].
\end{equation}
$D_{\mathrm{exact}}$ asks how often an entirely correct answer is corrupted,
which is the most direct reading of the invariance. Its denominator, however,
depends on the model, because different models solve different bases exactly.
Comparisons across models therefore use $D_{\mathrm{all}}$, and paired tests
restrict $D_{\mathrm{exact}}$ to the bases that both models solve. To compare
graph sizes we also report a per-edge rate, the fraction of all protected edges
whose prediction changes, which unlike the graph-level rates does not compound
over the protected edges of a graph.

For an answer-changing pair, the oracle identifies the original edges whose
labels change: one for the two-edge attachment, two for the in-base edit. The
\emph{response rate} is the fraction of all pairs on which the decoded
prediction for at least one of these edges differs from the prediction on $G$;
we call it the \emph{attachment response} for the two-edge attachment and the
\emph{in-base response} for the in-base edit.
A response requires only that the prediction change, and a change elsewhere in
the graph does not count. A response is \emph{correct} when the predictions on
all changed edges equal the new labels; correctness is reported on exactly
solved bases.

Because the in-base edits share base graphs with the bridge grafts, we can also
ask, on each exactly solved base, whether a model is both invariant to the graft
and correct on the in-base edit. We call the fraction of exactly solved bases on
which both hold the \emph{off-family joint rate}, since the in-base edit lies
outside the family of attachments that training uses. It is the joint measure an evaluation would report if the in-base
edit were its only sensitivity check.

For shortest paths, the protected outputs are the predecessors of original
vertices, and the answer-changing control is a two-edge attachment certified by
Dijkstra's algorithm \citep{dijkstra1959} to shorten the distance of at least one
original vertex.
Scoring uses the original vertices whose distance changes, at least one of which
always gets a new predecessor. A response is a change in the predicted
predecessor of at least one of them, and a correct response predicts the new
predecessors of all of them; both are scored on exactly solved bases.

\subsection{Statistical protocol}
\label{sec:stats}

Intervals for rates are Wilson score intervals \citep{wilson1927}, and intervals
for per-edge rates come from a bootstrap that resamples graphs. Paired
comparisons on a shared subset of exactly solved bases use McNemar's exact test
on the discordant pairs, with Holm's correction where a table contains a family
of tests. All models are scored on the same seeded two-edge pairs, so their
responses to this control are compared in the same way, over all pairs; the
per-pair responses come from a second scoring of the same checkpoints
(Appendix~\ref{app:mpnn-numerics}). A rate conditioned on exactly solved bases
needs enough of them to be meaningful; cells with fewer than $500$, a minimum
fixed before any model was trained, are reported on all pairs only.

\section{Experimental design}
\label{sec:design}

The experiments answer four questions, listed in Table~\ref{tab:design} with
the models and edits used for each. Configuration details are in
Appendix~\ref{app:details}.

\begin{table}[t]
\caption{Experimental design. MST cells use $10{,}000$ bridge-graft pairs,
$1{,}000$ two-edge attachment pairs and one in-base weight edit per bridge-graft
base where the base admits one. SSSP cells use $5{,}000$ rooted bridge-graft
pairs and a two-edge attachment control, with no in-base edit. Seeds are $14$,
$15$ and $16$ unless stated.}
\label{tab:design}
\centering
\small
\begin{tabular}{@{}p{0.2\textwidth}p{0.36\textwidth}p{0.36\textwidth}@{}}
\toprule
question & models & edits and sizes \\
\midrule
Do accurate models respect the invariance? (\S\ref{sec:res-violations}) &
reproduced GDT (3 seeds), shortest-path-bias transformer, MPNN, GDT trained at $n=64$ &
bridge grafts at $64{+}64$ and $16{+}16$; composites of $16$ vertices;
disjoint unions \\
What does invariance training buy and cost? (\S\ref{sec:res-training}) &
GDT: ERM; augmentation + symmetric KL, $\lambda\in\{0,1,10\}$; KL only and
ReLIC-form, $\lambda\in\{1,10\}$. MPNN: ERM; augmentation + symmetric KL,
$\lambda\in\{0,1,10\}$ &
bridge grafts, two-edge attachments and in-base edits at $16{+}16$;
all-pair rates at $64{+}64$ \\
Can structure give invariance without the cost? (\S\ref{sec:res-structure}) &
frozen weights of the models above &
inference-time masks and blockwise execution on the same pairs \\
Does it extend to a second task? (\S\ref{sec:res-sssp}) &
global and blockwise SSSP predictors, trained separately &
rooted bridge grafts and two-edge attachments, 24-vertex bases \\
\bottomrule
\end{tabular}
\end{table}

\subsection{Models}
\label{sec:models}

The primary backbone is the graph transformer of \citet{gdt2025} in its released
implementation (GDT): 16 layers, width 384, 16 heads, and relative random-walk
probabilities (RRWP) as positional encodings \citep{ma2023grit}; random-walk-based
encodings are a common choice in graph transformers \citep{rampasek2022gps}. It also has a global classification (CLS) token that every
token reads from and writes to. We train it on the authors' own MST task, $10^6$
graphs of 16 vertices. On 64-vertex test graphs the three reproduction seeds
reach a per-edge macro-F1 of \val{repro.f1min}--\val{repro.f1max}, within the
published $0.9604\pm0.0091$ (RRWP row of their Table~1, three seeds). A second transformer replaces the random-walk
encoding by a shortest-path-distance attention bias in the style of
\citet{ying2021graphormer}; we call it the shortest-path-bias transformer (SPD in
tables). The third model is a message-passing network (MPNN)
with GINE-style layers \citep{xu2019gin,hu2020pretrain} of the same depth and
width. It has no global token, no all-pairs attention and no global positional
encoding, so its receptive field equals its depth. Finally, a GDT trained on
64-vertex graphs serves as a size control. All models run in bf16 arithmetic,
which the released GDT implementation requires (Appendix~\ref{app:details}).

\subsection{Training comparison}
\label{sec:design-training}

All runs in the comparison use the same optimiser and $11{,}719$ steps of $128$
base graphs each, so they differ only in their objective. The invariance
objectives also process, for each base graph $G$, a bridge graft $G'$ of an
equally sized subgraph from the same generator, and compare predictions on the
edges of $G$. Hint-ReLIC adds disconnected subgraphs instead; we use bridge
grafts because they are the edit whose invariance we evaluate and the connected
edit closest in form to the two-edge control, and because a disconnected
addition cannot change the output of a message-passing network at all
(Section~\ref{sec:res-violations}). The
objectives are:
\begin{itemize}
\item \emph{ERM}: cross-entropy on unedited graphs.
\item \emph{Augmentation + symmetric KL}: cross-entropy on all edges of both $G$
and $G'$, plus $\lambda$ times the symmetric Kullback--Leibler divergence between the two
predicted label distributions on the original edges. At $\lambda=0$ this is
supervised augmentation alone.
\item \emph{KL only}: cross-entropy on $G$ only, plus the same penalty. The
grafted graph is never supervised, so this isolates the penalty.
\item \emph{ReLIC-form}: cross-entropy on $G$ only, plus $\lambda$ times a
contrastive invariance term on the edge representations that the decoder reads.
The representation of each original edge in $G$ should be closer to the same
edge in $G'$ than to any other original edge of the graph (an InfoNCE loss), and
a Kullback--Leibler term aligns the two views' similarity distributions
(temperature $0.1$ and weight $1$, as in \citet{bevilacqua2023hintrelic}).
\end{itemize}
The ReLIC-form objective has the form of Hint-ReLIC's loss but does not
reproduce Hint-ReLIC. It acts on final edge representations, because this
backbone is trained without intermediate algorithmic steps, and its negatives
are the original edges only, whereas Hint-ReLIC's include the added nodes. The
MPNN is trained with ERM and with augmentation + symmetric KL. None of these
objectives ever sees an answer-changing pair. All these runs, including the ERM
controls, are separate from the reproduced models of Section~\ref{sec:models}, which use batches of 256 (Appendix~\ref{app:details}), so the two
ERM-trained families differ slightly in their numbers.
Section~\ref{sec:res-violations} and Tables~\ref{tab:seeds}, \ref{tab:seeds-struct},
\ref{tab:strata} and \ref{tab:arms_gdt}--\ref{tab:size} use the reproduced
models; Section~\ref{sec:res-training} and Tables~\ref{tab:training},
\ref{tab:compiled} and \ref{tab:training-seeds}--\ref{tab:training-n64} use the
training-comparison checkpoints.

\subsection{Inference-time interventions}
\label{sec:design-interventions}

The structural experiments change only how a trained model is run, never its
weights. They rely on the block structure of the input. A vertex shared by two
blocks is an \emph{articulation vertex}, and a bridge graft leaves the blocks of
$G$ intact, adding only the blocks of $H$ and the bridge itself. Four
interventions restrict global channels of the transformer: cutting the
classification token in both directions; computing the random-walk encoding
within blocks; restricting attention to token pairs that share a block; and all
three together. Under the block-membership mask, an articulation vertex still
has a single token, which tokens of both its blocks can attend to, so across
layers this token can relay information from one block to the other.

\emph{Blockwise execution} removes this shared state. A depth-first search
extracts the biconnected edge blocks \citep{tarjan1972}. Each block becomes an
independent input with its own copies of its articulation vertices; its input
features and positional encodings are recomputed, and the same frozen network
scores it. Bridges belong to every spanning forest and are set symbolically, and
all other scores come from the network. The spanning-forest decoder factorises
over blocks, so for MST no information passes between blocks at any stage, and a
bridge graft changes neither the input nor the predictions of any original
block under exact arithmetic. For shortest paths,
blocks are processed in order from the root, and one scalar per articulation
vertex, the model's own predicted distance, is passed on
(Appendix~\ref{app:compiler}). Blockwise execution does not make the predictor
correct: membership of every non-bridge edge still depends on learned scores.

\section{Results}
\label{sec:results}

\subsection{Accurate models violate the invariance}
\label{sec:res-violations}

\paragraph{Violations at the backbone's test size}
Table~\ref{tab:seeds} reports the reproduced GDT on 64-vertex bases with
64-vertex grafts; 64 vertices is the size at which the backbone's accuracy was
published. Across three seeds, the model changes a protected label on
\val{head.seedmin}--\val{head.seedmax}\% of the bases it solves exactly. For
seed 14, the reference run in later comparisons, violations occur on
\val{head.defects} of \val{head.exact} exactly solved bases and on
\val{head.allpair}\% of all pairs, at a structured macro-F1 of \val{head.f1} on
the unedited bases.
The shortest-path-bias transformer changes a protected
label on \val{spd.defects} of \val{spd.exact} exactly solved bases
(\val{spd.rate}\%; Table~\ref{tab:arms_spd}).

\begin{table}[t]
\caption{Reproduced GDT under standard execution, $64$ attached to $64$
vertices, three training seeds. Violations on exactly solved bases with Wilson
95\% intervals, and over all pairs; ``max drift'' is the largest change of a
protected logit margin; response is to the two-edge attachment over all control
pairs. Interventions on the same models are in Table~\ref{tab:seeds-struct}.}
\label{tab:seeds}
\centering
\small
\begin{adjustbox}{max width=\textwidth}
\begin{tabular}{@{}rrrrlrrrr@{}}
\toprule
& \multicolumn{4}{c}{exactly solved bases} & & & & \\
\cmidrule(lr){2-5}
seed & bases & violations & \% & 95\% CI & all pairs \% & max drift & base macro-F1 & response \% \\
\midrule
14 & 1{,}522 & 945 & 62.09 & [59.62, 64.49] & 75.70 & 30.5 & 0.9815 & 73.2 \\
15 & 1{,}365 & 899 & 65.86 & [63.30, 68.33] & 77.04 & 31.8 & 0.9809 & 70.9 \\
16 & 794 & 423 & 53.27 & [49.80, 56.72] & 70.50 & 32.7 & 0.9775 & 69.6 \\%
 
\bottomrule
\end{tabular}
\end{adjustbox}
\end{table}

\paragraph{Size}
Each of these grafts doubles the input, and the edited graph has eight times as
many vertices as a training graph. Two controls separate the edit from this size
shift. First, we hold the edited graph at the training size of 16 vertices and
vary the split between base and graft (Table~\ref{tab:inrange}). The model still
changes protected labels on \val{inrange.min}--\val{inrange.max}\% of the bases
it solves exactly, and every confidence interval excludes zero. Second, a GDT
trained on 64-vertex graphs, for which the 128-vertex edited graph is only twice
the training size, changes a protected label on \val{n64train.rate}\% of
\val{n64train.exact} exactly solved bases, at an in-distribution structured
macro-F1 of \val{n64train.f1}. Violations therefore do not require leaving the
training size, but they are far more frequent under extrapolation: per
protected edge, the 16-vertex model violates the invariance \val{size.amp} times
as often at 64 vertices as at 16 (Appendix~\ref{app:size}).

\paragraph{Added material or the connecting edge?}
The disjoint union adds the same material without the bridge. For the GDT at
$64{+}64$, it leaves \val{union.gdt.allpair}\% of all pairs with a violation,
against \val{head.allpair}\% with the bridge, and \val{union.gdt.rate}\% of
exactly solved bases against \val{head.rate}\%. The shortest-path-bias transformer has \val{union.spd.defects}
violations on \val{union.spd.exact} exactly solved bases without the bridge and
\val{spd.defects} on \val{spd.exact} with it. For these two global models,
added material alone accounts for the violations at this size, and the bridge
adds little.

\paragraph{The message-passing network}
The MPNN, evaluated at $16{+}16$ vertices, behaves as its architecture predicts.
Without a connecting edge no message can pass between the parts, and it has no
violation under the disjoint union on \val{union.mpnn.exact} exactly solved bases
(none on \val{union.mpnn32.exact} in fp32 either). With the bridge it changes a
protected label on \val{mpnn.rate}\% of \val{mpnn.exact} exactly solved bases
(\val{mpnn32.bridge.rate}\% in fp32; Table~\ref{tab:arms_mpnn}), so local computation alone does not
enforce the invariance: messages flow across the bridge. At 64 vertices the MPNN
solves only \val{mpnn.n64exact} of $10{,}000$ bases exactly, too few for a
conditional rate (Section~\ref{sec:stats}).

\subsection{Invariance training suppresses responses to answer-changing attachments}
\label{sec:res-training}

\begin{figure}[t]
\centering
\includegraphics[width=0.8\textwidth]{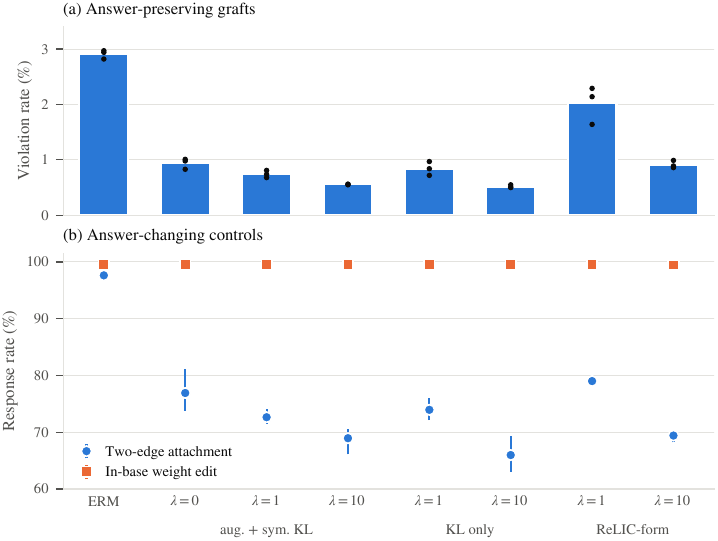}
\caption{GDT at $16{+}16$ vertices. \textbf{(a)} Violation rate: the fraction of
answer-preserving pairs on which any protected label changes; bars are
three-seed means and dots single seeds. \textbf{(b)} Response rate: the fraction
of answer-changing pairs on which the prediction changes on at least one edge
whose label the oracle changes, over all pairs; markers are three-seed means and
whiskers span the minimum and maximum over seeds (not confidence intervals).}
\label{fig:training}
\end{figure}

Table~\ref{tab:training} and Figure~\ref{fig:training} report the training
comparison on 16-vertex bases with 16-vertex attachments, where all models solve
most bases exactly. Each row averages three seeds; per-seed values, including
violations on exactly solved bases, are in Table~\ref{tab:training-seeds}. The
bases have the training size, but every edited graph has 32 vertices; the
invariance objectives were trained on such graphs and ERM was not.

\begin{table}[t]
\caption{Training comparison at $16{+}16$ vertices, three-seed means. Violation
and response rates are over all pairs. ``Attachment'' is the response to the
two-edge attachment and ``in-base edit'' the response to the in-base weight
edit. The off-family joint rate is the fraction of exactly solved bases on which
the model is invariant to the graft and correct on the in-base edit. ``ID F1'' is
structured macro-F1 on $1{,}000$ unedited 16-vertex graphs.}
\label{tab:training}
\centering
\small
\setlength{\tabcolsep}{3.5pt}
\begin{adjustbox}{max width=\textwidth}
\begin{tabular}{@{}llrrrrrr@{}}
\toprule
& & & violations & \multicolumn{2}{c}{response \%} & off-family & \\
\cmidrule(lr){5-6}
model & objective & $\lambda$ & \% & attachment & in-base edit & joint \% & ID F1 \\
\midrule
GDT & ERM & -- & 2.91 & 97.6 & 99.5 & 97.0 & 0.9984 \\
 & augmentation + sym.\ KL & 0 & 0.94 & 76.9 & 99.6 & 98.9 & 0.9987 \\
 & augmentation + sym.\ KL & 1 & 0.74 & 72.6 & 99.6 & 98.9 & 0.9985 \\
 & augmentation + sym.\ KL & 10 & 0.56 & 68.9 & 99.5 & 98.9 & 0.9985 \\
 & KL only & 1 & 0.84 & 73.9 & 99.6 & 98.8 & 0.9984 \\
 & KL only & 10 & 0.52 & 66.0 & 99.5 & 98.9 & 0.9981 \\
 & ReLIC-form & 1 & 2.02 & 79.0 & 99.5 & 97.6 & 0.9982 \\
 & ReLIC-form & 10 & 0.91 & 69.4 & 99.5 & 98.6 & 0.9982 \\
\midrule
MPNN & ERM & -- & 8.30 & 82.5 & 97.7 & 89.8 & 0.9909 \\
 & augmentation + sym.\ KL & 0 & 9.17 & 78.9 & 98.1 & 89.4 & 0.9920 \\
 & augmentation + sym.\ KL & 1 & 7.29 & 75.1 & 97.8 & 90.7 & 0.9913 \\
 & augmentation + sym.\ KL & 10 & 7.26 & 64.4 & 95.1 & 86.8 & 0.9795 \\%
 
\bottomrule
\end{tabular}
\end{adjustbox}
\end{table}

\paragraph{What training buys and what it costs}
We first describe the GDT; the MPNN follows below. Under ERM the GDT changes a
protected label on \val{r2.erm.allpair}\% of pairs. Every invariance objective
lowers this, to \val{r2.gdt.allpairmin}--\val{r2.gdt.allpairmax}\%
depending on the objective and $\lambda$, while structured macro-F1 on unedited
graphs stays between \val{r2.idf1.min} and \val{r2.idf1.max}. The same models
respond less often when an attached subgraph does change the answer: the
response to the two-edge attachment falls from \val{r2.erm.portal}\% of pairs
for ERM to \val{r2.gdt.portalmin}--\val{r2.gdt.portalmax}\% after training. The
loss is one-sided. On the same \val{r2.portal.pairs} pairs, ERM responds where
the trained model does not on \val{r2p.gdt.onlyerm.min}--\val{r2p.gdt.onlyerm.max}
pairs per model and seed, and the reverse happens on only
\val{r2p.gdt.onlytr.min}--\val{r2p.gdt.onlytr.max} (McNemar
$p<10^{\val{r2p.gdt.pexp}}$ after Holm's correction; Table~\ref{tab:paired-edit}).
On exactly solved bases a missing response has a concrete
meaning: the model keeps in the tree the one edge that the oracle removes
(Proposition~\ref{prop:oneflip}). There the response falls from
\val{r2.erm.portal_c}\% to \val{r2.lam10.portal_c}\% with augmentation and
$\lambda=10$.

\paragraph{The in-base response does not move}
The response to in-base weight edits stays near its ERM level of
\val{r2.erm.edit}\%, at \val{r2.gdt.editmin}--\val{r2.gdt.editmax}\% for every
invariance-trained GDT, including edits that lower the weight only slightly below
that of the tree edge it displaces (Appendix~\ref{app:edit-checks}). The rate
of correct responses is also similar: on the bases that a trained model and ERM of the
same seed both solve exactly, the number on which only ERM is correct on both
changed edges and the number on which only the trained model is never differ by
more than \val{r2.paired.maxshare}\% of the shared bases
(Table~\ref{tab:paired-edit}, beside the counts for the two-edge attachment).

\paragraph{Why an in-base check misses the cost}
Consider an evaluation that paired the invariance test with the in-base edit as
its only sensitivity check. Its joint measure, the off-family joint rate, rises
from \val{r2.erm.select}\% for ERM to \val{r2.lam10.select}\% after augmentation
with $\lambda=10$, and to \val{r2.klo10.select}\% with the penalty alone
(Table~\ref{tab:training}). Such an evaluation would report that
training improved invariance and preserved sensitivity. The loss appears only
under the control that attaches a subgraph, as the trained edits do.

\paragraph{The same trade-off across objectives}
The response loss appears with supervision alone, with the penalty alone, and
with a representation-level objective. Augmentation with $\lambda=0$ supervises
the grafted graph and applies no penalty; KL only applies the penalty and never
supervises the grafted graph. Both lower the attachment response, to
\val{r2.lam0.portal}\% and \val{r2.klo1.portal}\% (at $\lambda=1$)
respectively, so neither component is necessary for the loss. The ReLIC-form objective, which acts on representations
rather than outputs, shows the same pattern. At $\lambda=1$ it lowers violations
least, to \val{r2.rel1.allpair}\%, and still lowers the attachment response to
\val{r2.rel1.portal}\%.

\paragraph{The message-passing network}
In the MPNN, invariance training lowers the attachment response but reduces
violations far less than in the GDT. At $\lambda\le1$ the two responses separate as in the GDT,
though less sharply. With $\lambda=1$ the attachment response falls from
\val{r2.merm.portal}\% to \val{r2.mlam1.portal}\%, significantly for
\val{r2p.mlam1.nsig} seeds in the paired test, while the in-base response stays
at \val{r2.mlam1.edit}\% (ERM: \val{r2.merm.edit}\%). Augmentation alone lowers
the attachment response to \val{r2.mlam0.portal}\%, significantly for
\val{r2p.mlam0.nsig} of the three seeds, with the in-base response at
\val{r2.mlam0.edit}\%. Violations stay near the ERM
level of \val{r2.merm.allpair}\% of pairs, at \val{r2.mlam0.allpair}\% and
\val{r2.mlam1.allpair}\%. At $\lambda=10$ violations are
\val{r2.mlam10.allpair}\%, and the attachment response
(\val{r2.mlam10.portal}\%), the in-base response (\val{r2.mlam10.edit}\%) and
structured macro-F1 (\val{r2.merm.f1} to \val{r2.mlam10.f1}) all fall.

\paragraph{Size extrapolation}
At 64 vertices the GDT models trained at 16 solve few bases exactly after
invariance training (\val{r64.trained.exactmin}--\val{r64.trained.exactmax} of
$10{,}000$ per run), so we compare them on all pairs
(Table~\ref{tab:training-n64}). Invariance training lowers the fraction of pairs
with a violation from \val{r64.erm.allpair}\% for ERM to
\val{r64.trained.allpairmin}--\val{r64.trained.allpairmax}\%, but here both
responses fall. After augmentation with $\lambda=10$, for example, the attachment
response falls from
\val{r64.erm.portal}\% to \val{r64.lam10.portal}\% and the in-base response from
\val{r64.erm.edit}\% to \val{r64.lam10.edit}\%; Table~\ref{tab:training-n64}
gives every objective. Accuracy falls as well: in
per-edge macro-F1 on 64-vertex graphs, ERM scores \val{ood.erm.min}--\val{ood.erm.max} and the
invariance-trained models \val{ood.trained.min}--\val{ood.trained.max}. Under
size extrapolation the cost of invariance training is therefore visible without
any answer-changing control; on 16-vertex bases, as shown above, it is not.

\subsection{Blockwise execution as a structural reference}
\label{sec:res-structure}

The training results leave open whether the task itself forces a trade-off
between invariance and response. We examine this without retraining, by
changing only how a frozen model is run (Section~\ref{sec:design-interventions}).

\paragraph{Masks help at 64 vertices but not at the training size}
The first step is to restrict the global channels through which attached
material can reach protected outputs. On the frozen weights of the reproduced
GDT, restricting attention to token pairs that share a block lowers the all-pair
violation rate at $64{+}64$ from \val{head.allpair}\% to
\val{mask.gdt.attn.allpair}\% at unchanged structured macro-F1 (\val{head.f1}
and \val{mask.gdt.attn.f1}); for the shortest-path-bias transformer the rate
falls from \val{mask.spd.allpair}\% to \val{mask.spd.attn.allpair}\%
(Tables~\ref{tab:arms_gdt} and \ref{tab:arms_spd}). On the bases that all
interventions solve exactly, the attention mask repairs far more violations than it
introduces in every 64-vertex cell (Table~\ref{tab:paired}). Cutting the
classification token has significant effects but in inconsistent directions
across cells, and computing the positional encoding within blocks has no
significant effect at 64 vertices after Holm's correction. At the training size
the attention mask has the opposite effect. With an 8-vertex base and an 8-vertex graft,
which together have the 16 vertices of a training graph, it raises the all-pair
violation rate from \val{mask.inr.rel}\% to \val{mask.inr.attn}\%, and for the
GDT at $16{+}16$ it breaks more instances than it repairs. A mask removes some
cross-block dependence but also perturbs a computation the model learned
without it.

\begin{figure}[t]
\centering
\includegraphics[width=0.95\textwidth]{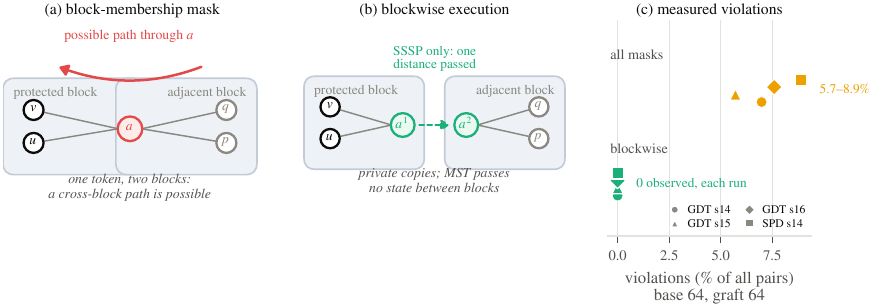}
\caption{Shared state under block-membership masking. \textbf{(a)} A mask that
allows all within-block attention leaves a possible path (arrow) through the
shared articulation token $a$. \textbf{(b)} Blockwise execution gives each block a private
copy. For MST no state passes between blocks; for source-rooted shortest paths
one scalar, the predicted distance of the articulation vertex, is passed in one
direction (dashed arrow). \textbf{(c)} Three GDT seeds and the
shortest-path-bias transformer at $64{+}64$ (marker shape identifies the run):
all three interventions together (``all masks'') leave \val{mask.allmin}--\val{mask.allmax}\% of pairs
with a violation, and blockwise execution none.}
\label{fig:mask}
\end{figure}

\paragraph{Why masks are not enough}
All three interventions together still leave \val{mask.gdt.all.allpair}\% (GDT)
and \val{mask.spd.all.allpair}\% (shortest-path bias) of pairs with a violation.
One reason is the shared token of the articulation vertex $a$
(Section~\ref{sec:design-interventions}; Figure~\ref{fig:mask}a): no mask that
keeps within-block attention can guarantee invariance for arbitrary weights. A unit test on a randomly initialised model shows that this path exists, though not that every measured violation uses it: the same masks
leave protected outputs exactly unchanged under a disjoint union, where no token
is shared, and change them under a bridge graft. A diagnostic mask that also
blocks attention from the protected side to the graft side lowers but does not
remove the violations (Appendix~\ref{app:asym}).

\paragraph{Blockwise execution removes the violations}
For both transformers, blockwise execution (Figure~\ref{fig:mask}b) gives zero
violations in every cell we evaluated, on exactly solved bases and on all pairs,
for every seed and checkpoint (Tables~\ref{tab:seeds-struct} and
\ref{tab:compiled}). Protected scores can still differ slightly under bf16
arithmetic, by at most \val{comp.drift.tf}, but no label changes. The MPNN has a few
exceptions, at most \val{r2c.mpnn.maxdef} exactly solved bases per cell,
consistent with rounding in its scatter operations
(Appendix~\ref{app:mpnn-numerics}).

\begin{table}[t]
\caption{Interventions on the reproduced GDT, $64$ attached to $64$ vertices,
three training seeds. Columns that compare executions use all pairs; ``exact
bases'' conditions on the model solving the base entirely, so its denominator
differs between executions. ``Masks'' is all three interventions together. Response is to
the two-edge attachment, over all control pairs.}
\label{tab:seeds-struct}
\centering
\small
\setlength{\tabcolsep}{3.5pt}
\begin{adjustbox}{max width=\textwidth}
\begin{tabular}{@{}lrrrrrrrrrr@{}}
\toprule
& \multicolumn{3}{c}{all pairs \%} & \multicolumn{3}{c}{exact bases}
& \multicolumn{2}{c}{base macro-F1} & \multicolumn{2}{c}{response \%} \\
\cmidrule(lr){2-4}\cmidrule(lr){5-7}\cmidrule(lr){8-9}\cmidrule(lr){10-11}
seed & standard & masks & blockw. & std.\ violations & \% & blockw.\ \% & std. & blockw. & std. & blockw. \\
\midrule
14 & 75.70 & 6.98 & \bfseries 0.00 & 945/1522 & 62.09 & \bfseries 0.00 & 0.9815 & 0.9817 & 73.2 & 72.8 \\
15 & 77.04 & 5.71 & \bfseries 0.00 & 899/1365 & 65.86 & \bfseries 0.00 & 0.9809 & 0.9810 & 70.9 & 70.8 \\
16 & 70.50 & 7.59 & \bfseries 0.00 & 423/794 & 53.27 & \bfseries 0.00 & 0.9775 & 0.9774 & 69.6 & 69.7 \\%
 
\bottomrule
\end{tabular}
\end{adjustbox}
\end{table}

\paragraph{Accuracy and response on the same weights}
The comparison that bears on the training results uses the checkpoints of
Section~\ref{sec:res-training}
(Table~\ref{tab:compiled}). On the ERM weights,
blockwise execution removes all violations (from \val{r2.erm.allpair}\% of pairs
to \val{r2c.erm.allpair}\%) and leaves the other measures where standard
execution had them: attachment response \val{r2c.erm.portal}\% against
\val{r2.erm.portal}\% (no seed differs in the paired test, $p\geq\val{r2p.blk.erm.pmin}$
after Holm's correction), in-base response \val{r2c.erm.edit}\% against
\val{r2.erm.edit}\%, and structured macro-F1 \val{r2c.erm.f1} against
\val{r2.erm.f1}. On these weights, then, blockwise execution reaches the
invariance that training did not, without the loss of response that training
incurred. On the augmentation + symmetric KL checkpoints it removes the
remaining violations and raises the attachment response by
\val{r2p.blk.aug.gainmin}--\val{r2p.blk.aug.gainmax} points, significantly on
every checkpoint ($p\leq\val{r2p.blk.aug.pmax}$ after Holm's correction), as it
does on the KL-only and ReLIC-form checkpoints. Most of the response lost in
training, however, remains under blockwise execution.

\begin{table}[t]
\caption{Blockwise execution on the checkpoints of the training comparison,
$16{+}16$ vertices, three-seed means. Rates over all pairs; ``exact bases'' is
the mean number of exactly solved bases out of $10{,}000$.}
\label{tab:compiled}
\centering
\small
\begin{adjustbox}{max width=\textwidth}
\begin{tabular}{@{}llrrrrr@{}}
\toprule
& & & violations & \multicolumn{2}{c}{response \%} & \\
\cmidrule(lr){5-6}
checkpoint & execution & exact bases & \% & attachment & in-base edit & ID F1 \\
\midrule
GDT ERM & standard & 9{,}813 & 2.91 & 97.6 & 99.5 & 0.9984 \\
 & blockwise & 9{,}811 & 0.00 & 98.1 & 99.5 & 0.9986 \\
GDT aug.\ + KL, $\lambda=0$ & standard & 9{,}800 & 0.94 & 76.9 & 99.6 & 0.9987 \\
 & blockwise & 9{,}769 & 0.00 & 82.1 & 99.5 & 0.9984 \\
GDT aug.\ + KL, $\lambda=1$ & standard & 9{,}795 & 0.74 & 72.6 & 99.6 & 0.9985 \\
 & blockwise & 9{,}760 & 0.00 & 79.6 & 99.5 & 0.9982 \\
GDT aug.\ + KL, $\lambda=10$ & standard & 9{,}790 & 0.56 & 68.9 & 99.5 & 0.9985 \\
 & blockwise & 9{,}737 & 0.00 & 73.4 & 99.4 & 0.9978 \\
GDT KL only, $\lambda=1$ & standard & 9{,}798 & 0.84 & 73.9 & 99.6 & 0.9984 \\
 & blockwise & 9{,}738 & 0.00 & 80.8 & 99.4 & 0.9979 \\
GDT KL only, $\lambda=10$ & standard & 9{,}797 & 0.52 & 66.0 & 99.5 & 0.9981 \\
 & blockwise & 9{,}757 & 0.00 & 74.8 & 99.4 & 0.9976 \\
GDT ReLIC-form, $\lambda=1$ & standard & 9{,}800 & 2.02 & 79.0 & 99.5 & 0.9982 \\
 & blockwise & 9{,}713 & 0.00 & 82.1 & 99.3 & 0.9974 \\
GDT ReLIC-form, $\lambda=10$ & standard & 9{,}785 & 0.91 & 69.4 & 99.5 & 0.9982 \\
 & blockwise & 9{,}599 & 0.00 & 76.3 & 99.1 & 0.9965 \\
MPNN ERM & standard & 8{,}910 & 8.30 & 82.5 & 97.7 & 0.9909 \\
 & blockwise & 8{,}971 & 0.01 & 84.1 & 97.8 & 0.9904 \\%
 
\bottomrule
\end{tabular}
\end{adjustbox}
\end{table}

\paragraph{Where blockwise execution is tested}
Retained accuracy is informative only where the bases themselves decompose,
which happens mainly at 16 vertices. The generator's 64-vertex bases are almost
always a single block (on average the largest block holds \val{blk.64.share}\% of
the vertices), so blockwise execution there amounts to scoring base and graft
separately. In 16-vertex bases the largest block holds \val{blk.16.share}\% of
the vertices on average, and at least $90\%$ in \val{blk.16.ge90}\% of bases.
Table~\ref{tab:strata} stratifies the reproduced seed-14 GDT by this share. Under
blockwise execution the fraction of bases solved exactly tracks standard
execution in every stratum (\val{blk.comp.exmin}--\val{blk.comp.exmax} against
\val{blk.rel.exmin}--\val{blk.rel.exmax}), whereas all three interventions together lower it
(\val{blk.mask.exmin}--\val{blk.mask.exmax}). Violations under standard execution
increase with the share of the largest block, from \val{blk.rel.lowdef} in the
lowest stratum to \val{blk.rel.highdef} (\val{blk.rel.highrate}\%) in the
highest; blockwise execution has none in any stratum.

\begin{table}[t]
\caption{Reproduced GDT, seed 14 (Section~\ref{sec:res-violations}), at
$16{+}16$ vertices, stratified by the share of base vertices in the base's
largest block. Exact rates are fractions of the bases in the stratum; violations
are counted on exactly solved bases. ``Masks'' is all three interventions
together.}
\label{tab:strata}
\centering
\small
\setlength{\tabcolsep}{4pt}
\begin{adjustbox}{max width=\textwidth}
\begin{tabular}{@{}lrrrrrrr@{}}
\toprule
& & \multicolumn{3}{c}{exactly solved (fraction)} & \multicolumn{3}{c}{violations / exact bases} \\
\cmidrule(lr){3-5}\cmidrule(lr){6-8}
largest block & bases & standard & blockwise & masks & standard & blockwise & masks \\
\midrule
$[0.0, 0.5)$ & 135 & 1.000 & 0.993 & 0.941 & 1/135 & 0/134 & 1/127 \\
$[0.5, 0.7)$ & 1{,}415 & 0.984 & 0.984 & 0.955 & 19/1{,}393 & 0/1{,}392 & 18/1{,}351 \\
$[0.7, 0.8)$ & 1{,}236 & 0.979 & 0.978 & 0.964 & 22/1{,}210 & 0/1{,}209 & 12/1{,}191 \\
$[0.8, 0.9)$ & 3{,}986 & 0.980 & 0.980 & 0.964 & 74/3{,}905 & 0/3{,}907 & 46/3{,}844 \\
$[0.9, 1.0]$ & 3{,}228 & 0.979 & 0.978 & 0.969 & 82/3{,}160 & 0/3{,}156 & 49/3{,}129 \\%
 
\bottomrule
\end{tabular}
\end{adjustbox}
\end{table}

\subsection{Shortest paths}
\label{sec:res-sssp}

The last experiment asks whether the structural reference extends to a second
task. Single-source shortest path decomposes like MST, except that one scalar
per articulation vertex, its distance from the root, must pass between blocks
(Proposition~\ref{prop:decomp}). Values flow only away from the root, and the
root lies in the original graph, so no original block receives anything from a
grafted block. Unlike the MST experiments, this one does not run frozen weights
in a new way: the global and blockwise predictors are trained separately, with
the same architecture and budget. On 24-vertex bases (Table~\ref{tab:rooted}),
the global predictor changes a provably fixed predecessor on
\val{sssp.globmin}--\val{sssp.globmax}\% of exactly solved bases, over the two
seeds whose training converged. The blockwise predictor has no violation on any
evaluated pair for any of three seeds. On answer-changing pairs whose base it
solves exactly, it responds on \val{sssp.respmin}--\val{sssp.respmax}\% and
predicts the correct predecessors of all vertices whose distance changes on
\val{sssp.corrmin}--\val{sssp.corrmax}\%. On shortest paths, too, a blockwise
predictor can be both invariant and responsive.

\begin{table}[t]
\caption{Source-rooted shortest path, 24-vertex bases. Response and correctness
are over answer-changing pairs whose base is solved exactly; correctness is not
conditional on responding. Seed 15's global predictor diverged in two training
attempts.}
\label{tab:rooted}
\centering
\small
\setlength{\tabcolsep}{4pt}
\begin{adjustbox}{max width=\textwidth}
\begin{tabular}{@{}lrrrrrrrr@{}}
\toprule
& \multicolumn{2}{c}{global, exact bases} & \multicolumn{2}{c}{blockwise, exact bases}
& \multicolumn{2}{c}{all pairs \%} & \multicolumn{2}{c}{blockwise response} \\
\cmidrule(lr){2-3}\cmidrule(lr){4-5}\cmidrule(lr){6-7}\cmidrule(lr){8-9}
seed & violations & \% & violations & \% & global & blockw. & response \% & correct \% \\
\midrule
14 & 290/4474 & 6.48 & \bfseries 0/4495 & \bfseries 0.00 & 8.90 & \bfseries 0.00 & 97.5 & 91.6 \\
15 & diverged & -- & \bfseries 0/4580 & \bfseries 0.00 & -- & \bfseries 0.00 & 98.0 & 94.7 \\
16 & 294/4449 & 6.61 & \bfseries 0/4406 & \bfseries 0.00 & 10.66 & \bfseries 0.00 & 93.6 & 86.8 \\%
 
\bottomrule
\end{tabular}
\end{adjustbox}
\end{table}

\section{Discussion}
\label{sec:discussion}

\paragraph{Matching the sensitivity check to the trained edits}
The main lesson concerns how invariance training is evaluated. Adding an
answer-changing test does not by itself expose a loss of response: our in-base
edits are certified and answer-changing, yet for the GDT on 16-vertex bases
they show no cost. A sensitivity
check for a
trained or claimed invariance should therefore include certified answer-changing
edits of the trained form. For augmentation-based invariance training in neural
algorithmic reasoning, this means testing on augmentations of the trained kind
that do change the answer, for example attachments through more than one edge
when training uses bridge or disconnected attachments.

\paragraph{What the trained models learned}
During training every attached subgraph preserved the answer, so responding
less to attached material is one way to satisfy the invariance objectives, and
nothing in them distinguishes attachments that would change the answer, since
none was ever shown. A model that takes this route also responds less to
answer-changing attachments. The GDT results are consistent with this across
output-level, representation-level and purely supervised objectives. We
did not identify the mechanism inside the network: features that mark the
attachment region, or attention that is reduced across new edges, would both
produce this behaviour. The result is a certified instance of the trade-off
described by \citet{tramer2020tradeoffs}, shown here for one kind of
answer-changing attachment.

\paragraph{Remedies}
One remedy, not tested here, is to train on certified answer-changing pairs of
the trained form alongside the answer-preserving ones, in the spirit of
counterfactual training \citep{kaushik2020cad,teney2020counterfactual}; whether
the learned sensitivity would then transfer to answer-changing edits of a kind
not seen in training is an open question. The structural route was tested here,
and its limit is one of scope. Blockwise execution shows that, on the GDT ERM
weights, the task permits invariance without loss of response, but it requires
the decomposition, which for these tasks is part of the algorithm, so it does
not carry over to tasks without such structure.

\paragraph{Limitations}
The two answer-changing controls differ in more than whether they attach
material (Section~\ref{sec:change}). The two-edge control uses light connecting
edges, is drawn by rejection and is not matched to the bridge-graft bases, and
the in-base edit lowers a weight rather than raising one. The data therefore
show that the choice of control changed the verdict here; they do not isolate
which of these differences is responsible. The comparison of responses rests on
16-vertex bases, since at 64 vertices few bases are solved exactly and both
responses fall. The study covers undirected graphs with positive, distinct
weights, and its answer-preserving edit is always a single-bridge graft.
Blockwise execution for extensions joined through two or more base vertices
would need richer interfaces between blocks and was not constructed; the
two-edge attachment joins at two vertices but serves only as an answer-changing
control. All instances come from one synthetic generator.

\section{Conclusion}
\label{sec:conclusion}

Graph algorithms make it possible to certify both that an edit cannot change an
answer and that another edit does. With such certificates we found that a
reproduced graph transformer changes provably fixed answers when material is
attached, and that on 16-vertex bases invariance training reduces these
violations at unchanged accuracy while lowering the model's response to
answer-changing attachments, a loss that a certified in-base edit does not
reveal. On the same ERM weights, blockwise execution
removes all observed violations while keeping accuracy and both responses, so
the task does not force the trade-off. Evaluations of trained or claimed
invariance should include certified answer-changing edits of the trained form.

\bibliographystyle{unsrtnat}
\bibliography{refs}

\begin{thebibliography}{43}
\providecommand{\natexlab}[1]{#1}
\providecommand{\url}[1]{\texttt{#1}}
\expandafter\ifx\csname urlstyle\endcsname\relax
  \providecommand{\doi}[1]{doi: #1}\else
  \providecommand{\doi}{doi: \begingroup \urlstyle{rm}\Url}\fi

\bibitem[Veli{\v{c}}kovi{\'c} et~al.(2020)Veli{\v{c}}kovi{\'c}, Ying, Padovano,
  Hadsell, and Blundell]{velickovic2020neural}
Petar Veli{\v{c}}kovi{\'c}, Rex Ying, Matilde Padovano, Raia Hadsell, and
  Charles Blundell.
\newblock Neural execution of graph algorithms.
\newblock In \emph{International Conference on Learning Representations
  (ICLR)}, 2020.

\bibitem[Veli{\v{c}}kovi{\'c} and Blundell(2021)]{velickovic2021nar}
Petar Veli{\v{c}}kovi{\'c} and Charles Blundell.
\newblock Neural algorithmic reasoning.
\newblock \emph{Patterns}, 2\penalty0 (7):\penalty0 100273, 2021.
\newblock \doi{10.1016/j.patter.2021.100273}.

\bibitem[Veli{\v{c}}kovi{\'c} et~al.(2022)Veli{\v{c}}kovi{\'c}, Badia, Budden,
  Pascanu, Banino, Dashevskiy, Hadsell, and Blundell]{velickovic2022clrs}
Petar Veli{\v{c}}kovi{\'c}, Adri{\`a}~Puigdom{\`e}nech Badia, David Budden,
  Razvan Pascanu, Andrea Banino, Misha Dashevskiy, Raia Hadsell, and Charles
  Blundell.
\newblock The {CLRS} algorithmic reasoning benchmark.
\newblock In \emph{International Conference on Machine Learning (ICML)}, volume
  162 of \emph{Proceedings of Machine Learning Research}, 2022.

\bibitem[Bevilacqua et~al.(2023)Bevilacqua, Nikiforou, Ibarz, Bica, Paganini,
  Blundell, Mitrovic, and Veli{\v{c}}kovi{\'c}]{bevilacqua2023hintrelic}
Beatrice Bevilacqua, Kyriacos Nikiforou, Borja Ibarz, Ioana Bica, Michela
  Paganini, Charles Blundell, Jovana Mitrovic, and Petar Veli{\v{c}}kovi{\'c}.
\newblock Neural algorithmic reasoning with causal regularisation.
\newblock In \emph{International Conference on Machine Learning (ICML)}, volume
  202 of \emph{Proceedings of Machine Learning Research}, 2023.

\bibitem[Rodionov and Prokhorenkova(2023)]{rodionov2023nohints}
Gleb Rodionov and Liudmila Prokhorenkova.
\newblock Neural algorithmic reasoning without intermediate supervision.
\newblock In \emph{Advances in Neural Information Processing Systems
  (NeurIPS)}, volume~36, pages 51663--51674, 2023.
\newblock \doi{10.52202/075280-2249}.

\bibitem[Tram{\`e}r et~al.(2020)Tram{\`e}r, Behrmann, Carlini, Papernot, and
  Jacobsen]{tramer2020tradeoffs}
Florian Tram{\`e}r, Jens Behrmann, Nicholas Carlini, Nicolas Papernot, and
  J{\"o}rn-Henrik Jacobsen.
\newblock Fundamental tradeoffs between invariance and sensitivity to
  adversarial perturbations.
\newblock In \emph{International Conference on Machine Learning (ICML)}, volume
  119 of \emph{Proceedings of Machine Learning Research}, 2020.

\bibitem[Niu and Bansal(2018)]{niu2018oversensitivity}
Tong Niu and Mohit Bansal.
\newblock Adversarial over-sensitivity and over-stability strategies for
  dialogue models.
\newblock In \emph{Conference on Computational Natural Language Learning
  (CoNLL)}, pages 486--496, 2018.
\newblock \doi{10.18653/v1/K18-1047}.

\bibitem[Ribeiro et~al.(2020)Ribeiro, Wu, Guestrin, and
  Singh]{ribeiro2020checklist}
Marco~Tulio Ribeiro, Tongshuang Wu, Carlos Guestrin, and Sameer Singh.
\newblock Beyond accuracy: Behavioral testing of {NLP} models with {CheckList}.
\newblock In \emph{Annual Meeting of the Association for Computational
  Linguistics (ACL)}, pages 4902--4912, 2020.
\newblock \doi{10.18653/v1/2020.acl-main.442}.

\bibitem[Gardner et~al.(2020)Gardner, Artzi, Basmov, Berant, Bogin, Chen,
  Dasigi, Dua, Elazar, Gottumukkala, Gupta, Hajishirzi, Ilharco, Khashabi, Lin,
  Liu, Liu, Mulcaire, Ning, Singh, Smith, Subramanian, Tsarfaty, Wallace,
  Zhang, and Zhou]{gardner2020contrast}
Matt Gardner, Yoav Artzi, Victoria Basmov, Jonathan Berant, Ben Bogin, Sihao
  Chen, Pradeep Dasigi, Dheeru Dua, Yanai Elazar, Ananth Gottumukkala, Nitish
  Gupta, Hannaneh Hajishirzi, Gabriel Ilharco, Daniel Khashabi, Kevin Lin,
  Jiangming Liu, Nelson~F. Liu, Phoebe Mulcaire, Qiang Ning, Sameer Singh,
  Noah~A. Smith, Sanjay Subramanian, Reut Tsarfaty, Eric Wallace, Ally Zhang,
  and Ben Zhou.
\newblock Evaluating models' local decision boundaries via contrast sets.
\newblock In \emph{Findings of the Association for Computational Linguistics:
  EMNLP 2020}, pages 1307--1323, 2020.
\newblock \doi{10.18653/v1/2020.findings-emnlp.117}.

\bibitem[Stoll et~al.(2025)Stoll, M{\"u}ller, and Morris]{gdt2025}
Timo Stoll, Luis M{\"u}ller, and Christopher Morris.
\newblock Generalizable insights for graph transformers in theory and practice.
\newblock In \emph{Advances in Neural Information Processing Systems
  (NeurIPS)}, volume~38, pages 187486--187534, 2025.
\newblock \doi{10.52202/085713-5637}.

\bibitem[Xu et~al.(2020)Xu, Li, Zhang, Du, Kawarabayashi, and
  Jegelka]{xu2020reason}
Keyulu Xu, Jingling Li, Mozhi Zhang, Simon~S. Du, Ken-ichi Kawarabayashi, and
  Stefanie Jegelka.
\newblock What can neural networks reason about?
\newblock In \emph{International Conference on Learning Representations
  (ICLR)}, 2020.

\bibitem[Dudzik and Veli{\v{c}}kovi{\'c}(2022)]{dudzik2022dp}
Andrew Dudzik and Petar Veli{\v{c}}kovi{\'c}.
\newblock Graph neural networks are dynamic programmers.
\newblock In \emph{Advances in Neural Information Processing Systems
  (NeurIPS)}, volume~35, pages 20635--20647, 2022.
\newblock \doi{10.52202/068431-1500}.

\bibitem[Xu et~al.(2021)Xu, Zhang, Li, Du, Kawarabayashi, and
  Jegelka]{xu2021extrapolate}
Keyulu Xu, Mozhi Zhang, Jingling Li, Simon~S. Du, Ken-ichi Kawarabayashi, and
  Stefanie Jegelka.
\newblock How neural networks extrapolate: From feedforward to graph neural
  networks.
\newblock In \emph{International Conference on Learning Representations
  (ICLR)}, 2021.

\bibitem[Mahdavi et~al.(2023)Mahdavi, Swersky, Kipf, Hashemi, Thrampoulidis,
  and Liao]{mahdavi2023ood}
Sadegh Mahdavi, Kevin Swersky, Thomas Kipf, Milad Hashemi, Christos
  Thrampoulidis, and Renjie Liao.
\newblock Towards better out-of-distribution generalization of neural
  algorithmic reasoning tasks.
\newblock \emph{Transactions on Machine Learning Research}, 2023.

\bibitem[Mitrovic et~al.(2021)Mitrovic, McWilliams, Walker, Buesing, and
  Blundell]{mitrovic2021relic}
Jovana Mitrovic, Brian McWilliams, Jacob Walker, Lars Buesing, and Charles
  Blundell.
\newblock Representation learning via invariant causal mechanisms.
\newblock In \emph{International Conference on Learning Representations
  (ICLR)}, 2021.

\bibitem[Bechler-Speicher et~al.(2024)Bechler-Speicher, Amos, Gilad-Bachrach,
  and Globerson]{bechlerspeicher2024graphs}
Maya Bechler-Speicher, Ido Amos, Ran Gilad-Bachrach, and Amir Globerson.
\newblock Graph neural networks use graphs when they shouldn't.
\newblock In \emph{International Conference on Machine Learning (ICML)}, volume
  235 of \emph{Proceedings of Machine Learning Research}, 2024.

\bibitem[Shi et~al.(2023)Shi, Chen, Misra, Scales, Dohan, Chi, Sch{\"a}rli, and
  Zhou]{shi2023distracted}
Freda Shi, Xinyun Chen, Kanishka Misra, Nathan Scales, David Dohan, Ed~H. Chi,
  Nathanael Sch{\"a}rli, and Denny Zhou.
\newblock Large language models can be easily distracted by irrelevant context.
\newblock In \emph{International Conference on Machine Learning (ICML)}, volume
  202 of \emph{Proceedings of Machine Learning Research}, pages 31210--31227,
  2023.

\bibitem[Yoran et~al.(2024)Yoran, Wolfson, Ram, and Berant]{yoran2024robust}
Ori Yoran, Tomer Wolfson, Ori Ram, and Jonathan Berant.
\newblock Making retrieval-augmented language models robust to irrelevant
  context.
\newblock In \emph{International Conference on Learning Representations
  (ICLR)}, 2024.

\bibitem[Arjovsky et~al.(2019)Arjovsky, Bottou, Gulrajani, and
  Lopez-Paz]{arjovsky2019irm}
Martin Arjovsky, L{\'e}on Bottou, Ishaan Gulrajani, and David Lopez-Paz.
\newblock Invariant risk minimization.
\newblock \emph{arXiv preprint arXiv:1907.02893}, 2019.
\newblock \doi{10.48550/arXiv.1907.02893}.

\bibitem[Jacobsen et~al.(2019)Jacobsen, Behrmann, Zemel, and
  Bethge]{jacobsen2019excessive}
J{\"o}rn-Henrik Jacobsen, Jens Behrmann, Richard Zemel, and Matthias Bethge.
\newblock Excessive invariance causes adversarial vulnerability.
\newblock In \emph{International Conference on Learning Representations
  (ICLR)}, 2019.

\bibitem[Benton et~al.(2020)Benton, Finzi, Izmailov, and
  Wilson]{benton2020augerino}
Gregory Benton, Marc Finzi, Pavel Izmailov, and Andrew~Gordon Wilson.
\newblock Learning invariances in neural networks from training data.
\newblock In \emph{Advances in Neural Information Processing Systems
  (NeurIPS)}, volume~33, pages 17605--17616, 2020.

\bibitem[Xiao et~al.(2021)Xiao, Wang, Efros, and Darrell]{xiao2021contrastive}
Tete Xiao, Xiaolong Wang, Alexei~A. Efros, and Trevor Darrell.
\newblock What should not be contrastive in contrastive learning.
\newblock In \emph{International Conference on Learning Representations
  (ICLR)}, 2021.

\bibitem[Kirichenko et~al.(2023)Kirichenko, Ibrahim, Balestriero, Bouchacourt,
  Vedantam, Firooz, and Wilson]{kirichenko2023classlevel}
Polina Kirichenko, Mark Ibrahim, Randall Balestriero, Diane Bouchacourt,
  Ramakrishna Vedantam, Hamed Firooz, and Andrew~Gordon Wilson.
\newblock Understanding the detrimental class-level effects of data
  augmentation.
\newblock In \emph{Advances in Neural Information Processing Systems
  (NeurIPS)}, volume~36, pages 17498--17526, 2023.
\newblock \doi{10.52202/075280-0768}.

\bibitem[Geirhos et~al.(2020)Geirhos, Jacobsen, Michaelis, Zemel, Brendel,
  Bethge, and Wichmann]{geirhos2020shortcut}
Robert Geirhos, J{\"o}rn-Henrik Jacobsen, Claudio Michaelis, Richard Zemel,
  Wieland Brendel, Matthias Bethge, and Felix~A. Wichmann.
\newblock Shortcut learning in deep neural networks.
\newblock \emph{Nature Machine Intelligence}, 2\penalty0 (11):\penalty0
  665--673, 2020.
\newblock \doi{10.1038/s42256-020-00257-z}.

\bibitem[Kaushik et~al.(2020)Kaushik, Hovy, and Lipton]{kaushik2020cad}
Divyansh Kaushik, Eduard Hovy, and Zachary~C. Lipton.
\newblock Learning the difference that makes a difference with
  counterfactually-augmented data.
\newblock In \emph{International Conference on Learning Representations
  (ICLR)}, 2020.

\bibitem[Teney et~al.(2020)Teney, Abbasnejad, and van~den
  Hengel]{teney2020counterfactual}
Damien Teney, Ehsan Abbasnejad, and Anton van~den Hengel.
\newblock Learning what makes a difference from counterfactual examples and
  gradient supervision.
\newblock In \emph{European Conference on Computer Vision (ECCV)}, pages
  580--599, 2020.
\newblock \doi{10.1007/978-3-030-58607-2_34}.

\bibitem[Veitch et~al.(2021)Veitch, D'Amour, Yadlowsky, and
  Eisenstein]{veitch2021counterfactual}
Victor Veitch, Alexander D'Amour, Steve Yadlowsky, and Jacob Eisenstein.
\newblock Counterfactual invariance to spurious correlations: Why and how to
  pass stress tests.
\newblock In \emph{Advances in Neural Information Processing Systems
  (NeurIPS)}, volume~34, 2021.

\bibitem[Zhang et~al.(2023)Zhang, Luo, Wang, and He]{zhang2023biconnectivity}
Bohang Zhang, Shengjie Luo, Liwei Wang, and Di~He.
\newblock Rethinking the expressive power of {GNN}s via graph biconnectivity.
\newblock In \emph{International Conference on Learning Representations
  (ICLR)}, 2023.

\bibitem[Dimitrov et~al.(2023)Dimitrov, Zhao, Abboud, and
  Ceylan]{dimitrov2023plane}
Radoslav Dimitrov, Zeyang Zhao, Ralph Abboud, and {\.I}smail~{\.I}lkan Ceylan.
\newblock {PlanE}: Representation learning over planar graphs.
\newblock In \emph{Advances in Neural Information Processing Systems
  (NeurIPS)}, volume~36, pages 16028--16054, 2023.
\newblock \doi{10.52202/075280-0705}.

\bibitem[Chen et~al.(2026)Chen, Mo, Ip, Zhang, Wu, Li, and
  U]{chen2026cosparsify}
Rongqin Chen, Fan Mo, Pak~Lon Ip, Shenghui Zhang, Dan Wu, Ye~Li, and Leong~Hou
  U.
\newblock Connectivity-guided sparsification of 2-{FWL} {GNN}s: Preserving full
  expressivity with improved efficiency.
\newblock In \emph{AAAI Conference on Artificial Intelligence (AAAI)},
  volume~40, pages 20226--20234, 2026.
\newblock \doi{10.1609/aaai.v40i24.39110}.

\bibitem[Li et~al.(2025)Li, Chen, Hopcroft, and He]{li2025damgt}
Chenyang Li, Jinsong Chen, John~E. Hopcroft, and Kun He.
\newblock {DAM-GT}: Dual positional encoding-based attention masking graph
  transformer for node classification.
\newblock \emph{arXiv preprint arXiv:2505.17660}, 2025.
\newblock \doi{10.48550/arXiv.2505.17660}.

\bibitem[Tarjan(1972)]{tarjan1972}
Robert Tarjan.
\newblock Depth-first search and linear graph algorithms.
\newblock \emph{SIAM Journal on Computing}, 1\penalty0 (2):\penalty0 146--160,
  1972.
\newblock \doi{10.1137/0201010}.

\bibitem[Li and Wang(2025)]{li2025agnncert}
Jiate Li and Binghui Wang.
\newblock {AGNNC}ert: Defending graph neural networks against arbitrary
  perturbations with deterministic certification.
\newblock In \emph{USENIX Security Symposium}, 2025.

\bibitem[Rodionov and Prokhorenkova(2025)]{rodionov2025discrete}
Gleb Rodionov and Liudmila Prokhorenkova.
\newblock Discrete neural algorithmic reasoning.
\newblock In \emph{International Conference on Machine Learning (ICML)}, volume
  267 of \emph{Proceedings of Machine Learning Research}, 2025.

\bibitem[Qharabagh et~al.(2026)Qharabagh, de~Luca, Giapitzakis, and
  Fountoulakis]{fetrat2026exact}
Muhammad~Fetrat Qharabagh, Artur~Back de~Luca, George Giapitzakis, and Kimon
  Fountoulakis.
\newblock Learning to execute graph algorithms exactly with graph neural
  networks.
\newblock \emph{arXiv preprint arXiv:2601.23207}, 2026.
\newblock \doi{10.48550/arXiv.2601.23207}.

\bibitem[Kruskal(1956)]{kruskal1956}
Joseph~B. Kruskal.
\newblock On the shortest spanning subtree of a graph and the traveling
  salesman problem.
\newblock \emph{Proceedings of the American Mathematical Society}, 7\penalty0
  (1):\penalty0 48--50, 1956.
\newblock \doi{10.1090/S0002-9939-1956-0078686-7}.

\bibitem[Dijkstra(1959)]{dijkstra1959}
Edsger~W. Dijkstra.
\newblock A note on two problems in connexion with graphs.
\newblock \emph{Numerische Mathematik}, 1:\penalty0 269--271, 1959.
\newblock \doi{10.1007/BF01386390}.

\bibitem[Wilson(1927)]{wilson1927}
Edwin~B. Wilson.
\newblock Probable inference, the law of succession, and statistical inference.
\newblock \emph{Journal of the American Statistical Association}, 22\penalty0
  (158):\penalty0 209--212, 1927.
\newblock \doi{10.1080/01621459.1927.10502953}.

\bibitem[Ma et~al.(2023)Ma, Lin, Lim, Romero-Soriano, Dokania, Coates, Torr,
  and Lim]{ma2023grit}
Liheng Ma, Chen Lin, Derek Lim, Adriana Romero-Soriano, Puneet~K. Dokania, Mark
  Coates, Philip Torr, and Ser-Nam Lim.
\newblock Graph inductive biases in transformers without message passing.
\newblock In \emph{International Conference on Machine Learning (ICML)}, volume
  202 of \emph{Proceedings of Machine Learning Research}, 2023.

\bibitem[Ramp{\'a}{\v{s}}ek et~al.(2022)Ramp{\'a}{\v{s}}ek, Galkin, Dwivedi,
  Luu, Wolf, and Beaini]{rampasek2022gps}
Ladislav Ramp{\'a}{\v{s}}ek, Michael Galkin, Vijay~Prakash Dwivedi, Anh~Tuan
  Luu, Guy Wolf, and Dominique Beaini.
\newblock Recipe for a general, powerful, scalable graph transformer.
\newblock In \emph{Advances in Neural Information Processing Systems
  (NeurIPS)}, volume~35, pages 14501--14515, 2022.
\newblock \doi{10.52202/068431-1054}.

\bibitem[Ying et~al.(2021)Ying, Cai, Luo, Zheng, Ke, He, Shen, and
  Liu]{ying2021graphormer}
Chengxuan Ying, Tianle Cai, Shengjie Luo, Shuxin Zheng, Guolin Ke, Di~He,
  Yanming Shen, and Tie-Yan Liu.
\newblock Do transformers really perform badly for graph representation?
\newblock In \emph{Advances in Neural Information Processing Systems
  (NeurIPS)}, volume~34, 2021.

\bibitem[Xu et~al.(2019)Xu, Hu, Leskovec, and Jegelka]{xu2019gin}
Keyulu Xu, Weihua Hu, Jure Leskovec, and Stefanie Jegelka.
\newblock How powerful are graph neural networks?
\newblock In \emph{International Conference on Learning Representations
  (ICLR)}, 2019.

\bibitem[Hu et~al.(2020)Hu, Liu, Gomes, Zitnik, Liang, Pande, and
  Leskovec]{hu2020pretrain}
Weihua Hu, Bowen Liu, Joseph Gomes, Marinka Zitnik, Percy Liang, Vijay Pande,
  and Jure Leskovec.
\newblock Strategies for pre-training graph neural networks.
\newblock In \emph{International Conference on Learning Representations
  (ICLR)}, 2020.

\end{thebibliography}

\appendix
\counterwithin{table}{section}

\section{Proofs}
\label{app:proofs}

\paragraph{Proposition~\ref{prop:scope}}
Let $G'$ be a bridge graft of $G$ with bridge $b=\{a,h\}$. Removing $b$
disconnects $V$ from $V_H$, so $b$ is a cut edge and lies on no cycle; every
cycle of $G'$ that meets both $V$ and $V_H$ would have to cross between them, and
the only crossing is $b$. Hence every cycle of $G'$ lies in $G$ or in $H$.

(i) With distinct weights the minimum spanning forest is unique
\citep{kruskal1956}, and by the cycle property an edge is excluded if and only if
it is the heaviest edge of some cycle. For $e\in E$ the cycles of $G'$ through $e$ are exactly the
cycles of $G$ through $e$, so $e$ has the same label in both.

(ii) Let the root $r$ lie in $V$. A walk from $r$ to $v\in V$ that visits $V_H$
crosses $b$ on the way out and again on the way back; deleting the part between
the two crossings gives a strictly shorter walk under positive weights. So no
shortest path to $v\in V$ enters $V_H$, $d_{G'}(v)=d_G(v)$, and the set of
shortest paths within $G$ is unchanged. Distinct weights do not make shortest
paths unique ($1+4=2+3$), so decoded predecessors use a fixed rule: among the
optimal predecessors take the smallest vertex identifier. A graft numbers its
vertices above all existing identifiers, so ties among original predecessors are
resolved identically before and after the graft.\qed

\paragraph{Proposition~\ref{prop:oneflip}}
Let $T_G$ and $T_H$ be the minimum spanning trees of the connected graphs $G$
and $H$, and let $G'=G\cup H\cup\{p_1,p_2\}$. An edge of $G$ outside $T_G$ is the
heaviest edge of some cycle of $G$, which is also a cycle of $G'$, so it is
outside the minimum spanning tree $T'$ of $G'$; the same holds for $H$. Hence
$T'\subseteq U=T_G\cup T_H\cup\{p_1,p_2\}$. $U$ has $(|V|-1)+(|V_H|-1)+2$ edges on
$|V|+|V_H|$ vertices and is connected, so it contains exactly one cycle $C$:
$p_1$, the path in $T_H$ from $h_1$ to $h_2$, $p_2$, and the path in $T_G$ from
$a_2$ back to $a_1$. A spanning tree of $G'$ inside $U$ must drop one edge of
$C$, and by the cycle property it drops the heaviest, so $T'=U\setminus\{m\}$
with $m$ the heaviest edge of $C$. If $m$ lies in $T_G$, exactly one original
edge changes label, from $1$ to $0$, and no original non-tree edge enters $T'$.
If $m$ is $p_1$, $p_2$ or an edge of $T_H$, no original label changes, and such a
draw is rejected by construction.\qed

\paragraph{Proposition~\ref{prop:exchange}}
Let $T$ be the minimum spanning tree of $G$, $e=\{u,v\}\notin T$, $P$ the path in
$T$ from $u$ to $v$, and $f$ the heaviest edge of $P$. Lower $w_e$ to a value
$w_e'<w_f$ and let $T^\ast=(T\setminus\{f\})\cup\{e\}$, a spanning tree. With
distinct weights, a spanning tree is the minimum one if and only if every edge
outside it is the heaviest edge on the cycle it closes with the tree. For $f$,
that cycle is $P\cup\{e\}$, and $f$ is its heaviest edge because the other edges
of $P$ are lighter than $f$ and $w_e'<w_f$. Take any other edge $g\notin T^\ast$,
so $g\notin T$ and $g\neq e$, and let $Q$ be the path in $T$ between its
endpoints; since $T$ is minimal, every edge of $Q$ is lighter than $g$. If $Q$
avoids $f$, it is also the path in $T^\ast$ and nothing changes. If $Q$ contains
$f$, then $w_f<w_g$, and the path in $T^\ast$ between the endpoints of $g$ uses
only edges of $Q\setminus\{f\}$, which are lighter than $g$, and edges of
$P\setminus\{f\}$ together with $e$, which are lighter than $w_f$ and hence than
$g$. So $g$ is again the heaviest edge of its cycle, $T^\ast$
is the minimum spanning tree after the edit, and exactly the labels of $e$ and
$f$ change. Each generated edit is also verified with an independent Kruskal
implementation.\qed

\begin{proposition}[Task decomposition]
\label{prop:decomp}
Minimum spanning forest membership of an edge depends only on the block that
contains it. For source-rooted shortest path, with the block-cut tree oriented
away from the root, $d(v)=d(a)+d_B(a,v)$ for every vertex $v$ of a block $B$
whose articulation vertex towards the root is $a$, where $d_B$ is the distance
within $B$.
\end{proposition}

\paragraph{Proof}
Every cycle through an edge lies in its block, which with the cycle property
gives the first claim. Every root-to-$v$ path enters $B$ through $a$, which gives
the second.\qed

\section{Blockwise execution}
\label{app:compiler}

A depth-first search returns the biconnected \emph{edge} blocks, so a vertex can
appear in several blocks and each edge belongs to exactly one. Each block becomes
an independent input: its vertices are relabelled to a local range, so an
articulation vertex that appears in $k$ blocks becomes $k$ tokens with no shared
state. No mask is applied inside a block. For MST every block is scored by the
same weights, in one batch across all blocks of all instances, and the scores
are written back to the original edges by identity. Bridges are coloops of the
graphic matroid, belong to every spanning forest, and are set symbolically. The
forest is decoded from the combined scores; the decoder factorises over blocks,
so protected decisions depend only on their own blocks.

For source-rooted tasks, blocks are visited in block-cut order from the source.
Each block has one entry vertex, the source or the articulation vertex towards
it, whose value is known when the block runs. The value passed on is the model's
own prediction for that vertex in the block above, not an oracle value, and it is
composed with the block's local prediction by the semiring product ($+$ for
shortest path). Errors therefore accumulate down the tree, and the reported
accuracy includes that accumulation. Single-edge blocks use the exact local
value.

\section{Experimental details}
\label{app:details}

\paragraph{Backbone and reproduction}
We use the released implementation of \citet{gdt2025} unmodified as the forward
model at its \texttt{PCQ} size, with \val{params.gdt} parameters. The released
code has $16$ layers, which we follow where the paper's text is inconsistent. The
published MST F1 of $0.9604\pm0.0091$ is reported by the paper as a
size-generalisation result; that it is macro-averaged over edges and measured on
64-vertex graphs we established from the released code. Our three reproduction
seeds score \val{repro.f1min}--\val{repro.f1max} on that test (\val{repro.f1} for
seed 14).

\paragraph{Training}
The reproduced GDT models use batch $256$, learning rate $3\times10^{-4}$, weight decay
$0.1$, dropout $0.1$, gradient clipping at $1.0$, $16$ random-walk steps, and
$11{,}719$ optimiser steps (the paper lists $11{,}718$) over $10^6$ training
graphs of 16 vertices, with
validation on $10{,}000$ graphs of 16 vertices and testing on $10{,}000$ graphs of
64. The training comparison uses the same settings with batch $128$ for every
objective, including the ERM control. The MPNN uses the same depth and width.

\paragraph{Data}
Graphs come from the backbone's released generator with its shipped
configuration: Erd\H{o}s--R\'enyi edges with probability $0.1$ plus a
component-linking step, and weights $\mathrm{round}(\mathcal{U}(0,10),6)$
resampled until all are distinct. The 16-vertex evaluation bases have
\val{blk.16.edges} edges, \val{blk.16.bridges} bridges and \val{blk.16.blocks}
nontrivial blocks (blocks with more than one edge) on average, against
\val{blk.64.blocks} nontrivial blocks at 64 vertices, so bridges already occur in
training data, although grafting still changes the input
distribution. Labels come from an independent union-find Kruskal implementation
and from Dijkstra's algorithm; neither uses model predictions or generator
bookkeeping.

\paragraph{Numerics}
The released forward pass requires bf16 autocast, so every GDT number is
produced in bf16; we report maximum protected score drift separately from
discrete violations.

\paragraph{Compute}
Each job used one H100 GPU. Training an MST model took between
\val{train.min.min} and \val{train.max.min} minutes, except the GDT trained on
64-vertex graphs, which took \val{train.n64.hours} hours.

\section{Additional results}
\label{app:extra}

\subsection{Asymmetric diagnostic mask}
\label{app:asym}

The asymmetric mask forbids tokens on the protected side, including the
articulation vertex $a$, from attending to tokens on the graft side, while graft-side
tokens may attend to $a$. It leaves the classification token in
place, so a two-step path through that token remains, and it computes positional
encodings on the whole composite, so a protected token's input can already depend
on the graft. It also needs to know which side of the bridge is protected, which
an MST instance does not specify, so it is a diagnostic and not a method. It
gives \val{asym.n64train.rate}\% exact-base violations on the model trained at
64 vertices (maximum drift \val{asym.n64train.drift}), \val{asym.s14n64.rate}\%
(\val{asym.s14n64.allpair}\% of all pairs) on the reproduced seed-14 model at
$64{+}64$, and \val{asym.s14n16.rate}\% (\val{asym.s14n16.allpair}\%) at
$16{+}16$.

\subsection{In-base edits close to the threshold}
\label{app:edit-checks}

The in-base edit lowers the weight of $e$ to a value drawn below $w_f$ (Section~\ref{sec:change}).
Small edits are no exception to the high response: among the third of edits
whose new weight lies closest below $w_f$, every GDT model in the training
comparison responds on at least \val{r2.gdt.lowterc}\% of exactly solved bases.
Table~\ref{tab:paired-edit} gives the paired comparison with ERM over all
exactly solved bases, for every objective and seed.

\subsection{Numerical exceptions in the MPNN}
\label{app:mpnn-numerics}

The MPNN's scatter operations are not bitwise deterministic. Under blockwise
execution it changes a protected label on \val{comp.mpnn16.allpair}\% of all
pairs at 16 vertices, none of them on an exactly solved base, and on
\val{comp.mpnn64.allpair}\% at 64 vertices, a cell reported on all pairs only
(Section~\ref{sec:stats}). In the training comparison, \val{r2c.mpnn.badcells} of the
\val{r2c.mpnn.cells} MPNN cells have between one and \val{r2c.mpnn.maxdef}
exact-base violations, with protected drift
\val{r2c.mpnn.driftmin}--\val{r2c.mpnn.driftmax}. Run in fp32, the MPNN of
Section~\ref{sec:res-violations} under blockwise execution has no violation on
its \val{comp.mpnn32.exact} exactly solved bases and
\val{comp.mpnn32.allpair}\% on all pairs at 16 vertices; the
training-comparison cells were not re-run in fp32. The same non-determinism
shows when the two-edge control is scored a second time to record per-pair
responses (Section~\ref{sec:stats}): every transformer cell reproduced its
original counts exactly, while \val{r2p.moved} of the \val{r2p.mpnncells} MPNN
cells moved by at most \val{r2p.moved.maxdiff} pairs. Rates quoted in the text
come from the first scoring and paired tests from the second.

\subsection{Size decomposition}
\label{app:size}

Per protected edge, GDT models trained and tested at the same size have similar
violation rates (\val{size.pe16}\% at 16 vertices and \val{size.pe64}\% at 64),
whereas the 16-vertex model tested at 64 vertices has \val{size.peood}\%,
\val{size.amp} times its rate at 16 vertices (Table~\ref{tab:size}). For the two
models tested at their training size, the graph-level rate grows with the number
of protected edges per graph, from \val{size.edges16} to \val{size.edges64}.

\subsection{Further tables}

\begin{table}[H]
\caption{Training comparison per seed, $16{+}16$ vertices. Violations over all
pairs and on exactly solved bases; responses over all pairs; ``correct'' is the
fraction of exactly solved bases on which both changed edges of the in-base
edit are predicted correctly.}
\label{tab:training-seeds}
\centering
\footnotesize
\setlength{\tabcolsep}{3pt}
\begin{adjustbox}{max width=\textwidth}
\begin{tabular}{@{}lllrrrrrrr@{}}
\toprule
& & & & \multicolumn{2}{c}{violations} & \multicolumn{2}{c}{response \%} & & \\
\cmidrule(lr){5-6}\cmidrule(lr){7-8}
model & objective & $\lambda$ & seed & all \% & exact & attach. & in-base & correct \% & ID F1 \\
\midrule
GDT & ERM & -- & 14 & 2.94 & 231/9{,}800 & 97.6 & 99.5 & 99.3 & 0.9984 \\
 &  &  & 15 & 2.97 & 226/9{,}816 & 97.8 & 99.5 & 99.2 & 0.9983 \\
 &  &  & 16 & 2.82 & 211/9{,}822 & 97.5 & 99.6 & 99.3 & 0.9984 \\
 & augmentation + sym.\ KL & 0 & 14 & 0.98 & 51/9{,}795 & 81.1 & 99.6 & 99.5 & 0.9987 \\
 &  &  & 15 & 0.83 & 47/9{,}803 & 75.8 & 99.6 & 99.4 & 0.9986 \\
 &  &  & 16 & 1.01 & 56/9{,}803 & 73.8 & 99.5 & 99.4 & 0.9988 \\
 & augmentation + sym.\ KL & 1 & 14 & 0.74 & 39/9{,}786 & 71.5 & 99.6 & 99.4 & 0.9981 \\
 &  &  & 15 & 0.81 & 44/9{,}796 & 72.4 & 99.6 & 99.3 & 0.9987 \\
 &  &  & 16 & 0.68 & 44/9{,}803 & 74.0 & 99.6 & 99.3 & 0.9987 \\
 & augmentation + sym.\ KL & 10 & 14 & 0.56 & 33/9{,}774 & 66.1 & 99.4 & 99.0 & 0.9984 \\
 &  &  & 15 & 0.55 & 33/9{,}798 & 70.2 & 99.6 & 99.3 & 0.9987 \\
 &  &  & 16 & 0.56 & 32/9{,}797 & 70.5 & 99.6 & 99.3 & 0.9982 \\
 & KL only & 1 & 14 & 0.72 & 44/9{,}802 & 73.6 & 99.6 & 99.3 & 0.9983 \\
 &  &  & 15 & 0.84 & 50/9{,}796 & 76.1 & 99.6 & 99.3 & 0.9984 \\
 &  &  & 16 & 0.97 & 69/9{,}796 & 72.1 & 99.6 & 99.4 & 0.9983 \\
 & KL only & 10 & 14 & 0.51 & 29/9{,}807 & 65.6 & 99.5 & 99.2 & 0.9977 \\
 &  &  & 15 & 0.50 & 28/9{,}778 & 63.0 & 99.6 & 99.2 & 0.9982 \\
 &  &  & 16 & 0.55 & 39/9{,}806 & 69.3 & 99.5 & 99.2 & 0.9983 \\
 & ReLIC-form & 1 & 14 & 2.14 & 188/9{,}802 & 78.9 & 99.5 & 99.4 & 0.9981 \\
 &  &  & 15 & 2.29 & 200/9{,}799 & 79.7 & 99.5 & 99.2 & 0.9979 \\
 &  &  & 16 & 1.64 & 131/9{,}798 & 78.4 & 99.5 & 99.4 & 0.9986 \\
 & ReLIC-form & 10 & 14 & 0.86 & 60/9{,}778 & 70.2 & 99.4 & 99.2 & 0.9983 \\
 &  &  & 15 & 0.99 & 65/9{,}787 & 69.8 & 99.5 & 99.2 & 0.9978 \\
 &  &  & 16 & 0.89 & 66/9{,}789 & 68.2 & 99.5 & 99.3 & 0.9983 \\
MPNN & ERM & -- & 14 & 8.20 & 528/8{,}908 & 83.2 & 97.7 & 95.5 & 0.9899 \\
 &  &  & 15 & 8.50 & 557/8{,}924 & 82.6 & 97.6 & 95.6 & 0.9916 \\
 &  &  & 16 & 8.19 & 538/8{,}899 & 81.7 & 97.7 & 95.7 & 0.9912 \\
 & augmentation + sym.\ KL & 0 & 14 & 9.22 & 657/9{,}064 & 78.2 & 98.1 & 96.3 & 0.9917 \\
 &  &  & 15 & 9.16 & 647/9{,}025 & 79.6 & 98.1 & 96.5 & 0.9926 \\
 &  &  & 16 & 9.14 & 659/9{,}050 & 79.0 & 98.0 & 96.4 & 0.9915 \\
 & augmentation + sym.\ KL & 1 & 14 & 7.33 & 474/8{,}895 & 74.7 & 97.7 & 95.6 & 0.9905 \\
 &  &  & 15 & 7.29 & 493/8{,}951 & 74.9 & 97.9 & 96.0 & 0.9918 \\
 &  &  & 16 & 7.24 & 474/8{,}957 & 75.8 & 97.9 & 95.9 & 0.9916 \\
 & augmentation + sym.\ KL & 10 & 14 & 8.06 & 427/7{,}742 & 61.0 & 95.2 & 91.2 & 0.9796 \\
 &  &  & 15 & 6.45 & 369/7{,}900 & 66.1 & 95.5 & 92.3 & 0.9815 \\
 &  &  & 16 & 7.28 & 369/7{,}487 & 66.0 & 94.7 & 90.8 & 0.9773 \\%
 
\bottomrule
\end{tabular}
\end{adjustbox}
\end{table}

\begin{table}[H]
\caption{Each trained GDT against ERM of the same seed, paired. In-base edit:
on the bases both solve exactly, the number of bases on which only ERM is
correct on both changed edges / only the trained model is. Two-edge attachment:
over all $1{,}000$ pairs, the number on which only ERM responds / only the
trained model does. $p$ is McNemar's exact test, uncorrected; after Holm's
correction over the 21 comparisons, every two-edge $p$ stays below
$10^{\val{r2p.gdt.pexp}}$, and the only in-base difference that remains
significant (seed 14, $\lambda=0$) favours the trained model.}
\label{tab:paired-edit}
\centering
\footnotesize
\setlength{\tabcolsep}{3pt}
\begin{adjustbox}{max width=\textwidth}
\begin{tabular}{@{}lrrrrrrrrrrrr@{}}
\toprule
& \multicolumn{4}{c}{seed 14} & \multicolumn{4}{c}{seed 15} & \multicolumn{4}{c}{seed 16} \\
\cmidrule(lr){2-5}\cmidrule(lr){6-9}\cmidrule(lr){10-13}
objective & in-base & $p$ & two-edge & $p$ & in-base & $p$ & two-edge & $p$
& in-base & $p$ & two-edge & $p$ \\
\midrule
sym.\ KL, $\lambda=0$ & 9/32 & $<$0.001 & 179/14 & $<$0.001 & 18/39 & 0.008 & 231/11 & $<$0.001 & 17/28 & 0.135 & 249/12 & $<$0.001 \\
sym.\ KL, $\lambda=1$ & 24/33 & 0.289 & 276/15 & $<$0.001 & 28/40 & 0.182 & 261/7 & $<$0.001 & 27/31 & 0.694 & 249/14 & $<$0.001 \\
sym.\ KL, $\lambda=10$ & 50/28 & 0.017 & 326/11 & $<$0.001 & 26/38 & 0.169 & 286/10 & $<$0.001 & 32/32 & 1.000 & 283/13 & $<$0.001 \\
KL only, $\lambda=1$ & 24/27 & 0.780 & 249/9 & $<$0.001 & 28/40 & 0.182 & 225/8 & $<$0.001 & 23/31 & 0.341 & 267/13 & $<$0.001 \\
KL only, $\lambda=10$ & 45/35 & 0.314 & 332/12 & $<$0.001 & 25/29 & 0.683 & 354/6 & $<$0.001 & 35/27 & 0.374 & 292/10 & $<$0.001 \\
ReLIC-form, $\lambda=1$ & 33/43 & 0.302 & 203/16 & $<$0.001 & 32/33 & 1.000 & 188/7 & $<$0.001 & 25/37 & 0.162 & 204/13 & $<$0.001 \\
ReLIC-form, $\lambda=10$ & 43/39 & 0.741 & 288/14 & $<$0.001 & 41/39 & 0.911 & 290/10 & $<$0.001 & 39/42 & 0.824 & 304/11 & $<$0.001 \\%
 
\bottomrule
\end{tabular}
\end{adjustbox}
\end{table}

\begin{table}[H]
\caption{Training comparison at $64{+}64$ vertices, three-seed means over all
pairs. Exact bases are too few for conditional rates. The last column is
per-edge macro-F1 on $10{,}000$ unedited 64-vertex graphs under the backbone's
released protocol, range over seeds.}
\label{tab:training-n64}
\centering
\small
\begin{adjustbox}{max width=\textwidth}
\begin{tabular}{@{}lrrrrrr@{}}
\toprule
& & & violations & \multicolumn{2}{c}{response \%} & size-OOD \\
\cmidrule(lr){5-6}
objective & $\lambda$ & exact bases & \% & attachment & in-base edit & macro-F1 \\
\midrule
ERM & -- & 1{,}046 & 76.2 & 69.7 & 95.3 & 0.960--0.964 \\
augmentation + sym.\ KL & 0 & 41 & 61.3 & 49.6 & 89.8 & 0.936--0.942 \\
augmentation + sym.\ KL & 1 & 32 & 55.4 & 49.5 & 88.7 & 0.925--0.937 \\
augmentation + sym.\ KL & 10 & 5 & 65.4 & 33.9 & 84.0 & 0.761--0.890 \\
KL only & 1 & 10 & 61.6 & 45.1 & 86.8 & 0.922--0.932 \\
KL only & 10 & 7 & 61.5 & 40.0 & 87.8 & 0.889--0.937 \\
ReLIC-form & 1 & 20 & 58.1 & 54.6 & 87.9 & 0.843--0.874 \\
ReLIC-form & 10 & 47 & 53.8 & 54.0 & 88.8 & 0.863--0.900 \\%
 
\bottomrule
\end{tabular}
\end{adjustbox}
\end{table}

\begin{table}[H]
\caption{Inference-time interventions, reproduced GDT, seed 14, at $64{+}64$. Exact-base
columns have arm-specific denominators; all-pair columns compare arms. Response
is to the two-edge attachment.}
\label{tab:arms_gdt}
\centering
\small
\setlength{\tabcolsep}{3pt}
\begin{adjustbox}{max width=\textwidth}
\begin{tabular}{@{}lrrlrrrr@{}}
\toprule
& \multicolumn{3}{c}{exact bases only} & \multicolumn{4}{c}{all pairs} \\
\cmidrule(lr){2-4}\cmidrule(lr){5-8}
arm & violations & \% & 95\% CI & violations \% & max drift & response \% & base F1 \\
\midrule
standard execution & 945/1522 & 62.09 & [59.62, 64.49] & 75.70 & 30.469 & 73.2 & 0.9815 \\
\quad cut CLS token & 770/1204 & 63.95 & [61.20, 66.62] & 77.90 & 30.969 & 72.8 & 0.9798 \\
\quad block-local RRWP & 939/1507 & 62.31 & [59.83, 64.72] & 75.57 & 29.312 & 73.3 & 0.9816 \\
\quad block-masked attention & 95/1543 & 6.16 & [5.06, 7.47] & 11.07 & 30.031 & 72.9 & 0.9816 \\
\quad all three masks & 35/1238 & 2.83 & [2.04, 3.91] & 6.98 & 28.094 & 72.8 & 0.9801 \\
\textbf{blockwise execution} & \bfseries 0/1543 & \bfseries 0.00 & [0.00, 0.25] & \bfseries 0.00 & 0.062 & 72.8 & 0.9817 \\%
 
\bottomrule
\end{tabular}
\end{adjustbox}
\end{table}

\begin{table}[H]
\caption{The same interventions on the shortest-path-bias transformer at
$64{+}64$. This backbone has no random-walk encoding to localise, so that row
repeats the standard-execution row.}
\label{tab:arms_spd}
\centering
\small
\setlength{\tabcolsep}{3pt}
\begin{adjustbox}{max width=\textwidth}
\begin{tabular}{@{}lrrlrrrr@{}}
\toprule
& \multicolumn{3}{c}{exact bases only} & \multicolumn{4}{c}{all pairs} \\
\cmidrule(lr){2-4}\cmidrule(lr){5-8}
arm & violations & \% & 95\% CI & violations \% & max drift & response \% & base F1 \\
\midrule
standard execution & 1013/1338 & 75.71 & [73.34, 77.93] & 83.99 & 29.500 & 70.6 & 0.9806 \\
\quad cut CLS token & 702/894 & 78.52 & [75.71, 81.09] & 86.46 & 29.844 & 67.0 & 0.9769 \\
\quad block-local RRWP & 1013/1338 & 75.71 & [73.34, 77.93] & 83.99 & 29.500 & 70.6 & 0.9806 \\
\quad block-masked attention & 80/1382 & 5.79 & [4.68, 7.15] & 14.18 & 26.969 & 70.9 & 0.9807 \\
\quad all three masks & 42/939 & 4.47 & [3.33, 5.99] & 8.89 & 19.234 & 66.7 & 0.9773 \\
\textbf{blockwise execution} & \bfseries 0/1389 & \bfseries 0.00 & [0.00, 0.28] & \bfseries 0.00 & 0.070 & 70.4 & 0.9807 \\%
 
\bottomrule
\end{tabular}
\end{adjustbox}
\end{table}

\begin{table}[H]
\caption{The same interventions on the MPNN of Section~\ref{sec:res-violations}
(not the training-comparison MPNN) at $16{+}16$. It has no
classification token, no global positional encoding and no all-pairs attention,
so the first three interventions have little to act on.}
\label{tab:arms_mpnn}
\centering
\small
\setlength{\tabcolsep}{3pt}
\begin{adjustbox}{max width=\textwidth}
\begin{tabular}{@{}lrrlrrrr@{}}
\toprule
& \multicolumn{3}{c}{exact bases only} & \multicolumn{4}{c}{all pairs} \\
\cmidrule(lr){2-4}\cmidrule(lr){5-8}
arm & violations & \% & 95\% CI & violations \% & max drift & response \% & base F1 \\
\midrule
standard execution & 552/9110 & 6.06 & [5.59, 6.57] & 7.98 & 40.000 & 84.6 & 0.9921 \\
\quad cut CLS token & 554/9111 & 6.08 & [5.61, 6.59] & 8.00 & 40.000 & 84.7 & 0.9921 \\
\quad block-local RRWP & 552/9111 & 6.06 & [5.59, 6.57] & 7.98 & 40.000 & 84.7 & 0.9921 \\
\quad block-masked attention & 534/9125 & 5.85 & [5.39, 6.35] & 7.67 & 39.125 & 84.6 & 0.9922 \\
\quad all three masks & 533/9122 & 5.84 & [5.38, 6.34] & 7.69 & 34.375 & 84.5 & 0.9922 \\
\textbf{blockwise execution} & \bfseries 0/9164 & \bfseries 0.00 & [0.00, 0.04] & \bfseries 0.01 & 0.125 & 86.2 & 0.9919 \\%
 
\bottomrule
\end{tabular}
\end{adjustbox}
\end{table}

\begin{table}[H]
\caption{Interventions on the reproduced models, compared on the bases that every
arm solves exactly.
Entries are violations removed / violations introduced on that subset, against
standard execution; $^{\ast}$ marks $p<0.05$ and $^{\ast\ast}$ $p<10^{-3}$ under
McNemar's exact test, uncorrected. Under Holm's correction over the table's
tests, every $^{\ast\ast}$ entry and the cut-CLS entry for GDT s14 remain
significant, and the other $^{\ast}$ entries do not.}
\label{tab:paired}
\centering
\small
\setlength{\tabcolsep}{5pt}
\begin{adjustbox}{max width=\textwidth}
\begin{tabular}{@{}lrrrrrr@{}}
\toprule
cell & shared & cut CLS & block RRWP & block attn. & all masks & blockwise \\
\midrule
GDT s14 & 808 & 65/105$^{\ast}$ & 12/17 & 408/10$^{\ast\ast}$ & 413/4$^{\ast\ast}$ & 423/0$^{\ast\ast}$ \\
GDT s15 & 1019 & 86/34$^{\ast\ast}$ & 14/12 & 603/6$^{\ast\ast}$ & 618/5$^{\ast\ast}$ & 635/0$^{\ast\ast}$ \\
GDT s16 & 418 & 91/35$^{\ast\ast}$ & 8/1$^{\ast}$ & 181/5$^{\ast\ast}$ & 189/5$^{\ast\ast}$ & 196/0$^{\ast\ast}$ \\
SPD s14 & 725 & 35/98$^{\ast\ast}$ & 0/0 & 478/0$^{\ast\ast}$ & 475/3$^{\ast\ast}$ & 496/0$^{\ast\ast}$ \\
GDT s14, $n{=}16$ & 9442 & 69/163$^{\ast\ast}$ & 52/144$^{\ast\ast}$ & 136/205$^{\ast\ast}$ & 141/104$^{\ast}$ & 149/0$^{\ast\ast}$ \\
MPNN s14, $n{=}16$ & 9019 & 1/2 & 1/0 & 92/62$^{\ast}$ & 92/63$^{\ast}$ & 525/0$^{\ast\ast}$ \\%
 
\bottomrule
\end{tabular}
\end{adjustbox}
\end{table}

\begin{table}[H]
\caption{Size-controlled cells, reproduced GDT, seed 14: the composite has
exactly the training size of 16 vertices and the split between base and graft varies. Base size and graph
structure still differ from training data.}
\label{tab:inrange}
\centering
\small
\setlength{\tabcolsep}{4pt}
\begin{adjustbox}{max width=\textwidth}
\begin{tabular}{@{}rrrlrrlrr@{}}
\toprule
base & graft & composite & arm & violations & \% & 95\% CI & all pairs \% & base F1 \\
\midrule
14 & 2 & 16 & standard execution & 26/9833 & 0.26 & [0.18, 0.39] & 0.45 & 0.9981 \\
14 & 2 & 16 & \quad block-masked attention & 259/9777 & 2.65 & [2.35, 2.99] & 2.78 & 0.9978 \\
14 & 2 & 16 & \quad block-local RRWP & 28/9824 & 0.29 & [0.20, 0.41] & 0.55 & 0.9979 \\
14 & 2 & 16 & \textbf{blockwise execution} & \bfseries 0/9827 & \bfseries 0.00 & [0.00, 0.04] & 0.00 & 0.9981 \\
14 & 2 & 16 & \quad cut CLS token & 28/9758 & 0.29 & [0.20, 0.41] & 0.67 & 0.9976 \\
14 & 2 & 16 & \quad all three masks & 79/9684 & 0.82 & [0.66, 1.02] & 1.11 & 0.9972 \\
12 & 4 & 16 & standard execution & 44/9848 & 0.45 & [0.33, 0.60] & 0.96 & 0.9986 \\
12 & 4 & 16 & \quad block-masked attention & 267/9823 & 2.72 & [2.41, 3.06] & 2.96 & 0.9983 \\
12 & 4 & 16 & \quad block-local RRWP & 46/9844 & 0.47 & [0.35, 0.62] & 0.95 & 0.9986 \\
12 & 4 & 16 & \textbf{blockwise execution} & \bfseries 0/9852 & \bfseries 0.00 & [0.00, 0.04] & 0.00 & 0.9985 \\
12 & 4 & 16 & \quad cut CLS token & 60/9789 & 0.61 & [0.48, 0.79] & 1.13 & 0.9984 \\
12 & 4 & 16 & \quad all three masks & 134/9760 & 1.37 & [1.16, 1.62] & 1.60 & 0.9981 \\
8 & 8 & 16 & standard execution & 45/9913 & 0.45 & [0.34, 0.61] & 0.87 & 0.9986 \\
8 & 8 & 16 & \quad block-masked attention & 185/9910 & 1.87 & [1.62, 2.15] & 2.02 & 0.9984 \\
8 & 8 & 16 & \quad block-local RRWP & 51/9916 & 0.51 & [0.39, 0.68] & 0.89 & 0.9986 \\
8 & 8 & 16 & \textbf{blockwise execution} & \bfseries 0/9908 & \bfseries 0.00 & [0.00, 0.04] & 0.00 & 0.9986 \\
8 & 8 & 16 & \quad cut CLS token & 38/9861 & 0.39 & [0.28, 0.53] & 1.05 & 0.9974 \\
8 & 8 & 16 & \quad all three masks & 87/9846 & 0.88 & [0.72, 1.09] & 0.98 & 0.9970 \\%
 
\bottomrule
\end{tabular}
\end{adjustbox}
\end{table}

\begin{table}[H]
\caption{Size decomposition, reproduced GDT models. Models trained at their base
size have similar per-edge violation rates; the 16-vertex model evaluated at 64
vertices has a higher one. In every row the graft doubles the base. ``Graph \%''
is the fraction of all pairs with a protected change; macro-F1 is structured
macro-F1 on unedited graphs of the evaluated size.}
\label{tab:size}
\centering
\small
\begin{adjustbox}{max width=\textwidth}
\begin{tabular}{@{}llrrrr@{}}
\toprule
$n$ & training & edges/graph & per-edge \% & graph \% & macro-F1 \\
\midrule
16 & trained at 16 & 24.8 & 0.229 & 2.81 & 0.9987 \\
32 & trained at 16 (size-OOD) & 54.5 & 0.849 & 21.56 & 0.9968 \\
64 & trained at 16 (size-OOD) & 198.3 & 1.263 & 75.70 & 0.9815 \\
64 & \textbf{trained at 64} & 201.9 & 0.202 & 18.69 & 0.9987 \\%
 
\bottomrule
\end{tabular}
\end{adjustbox}
\end{table}

\end{document}